\documentclass[sigconf]{acmart}
\AtBeginDocument{%
  \providecommand\BibTeX{{%
    \normalfont B\kern-0.5em{\scshape i\kern-0.25em b}\kern-0.8em\TeX}}}

\usepackage{microtype}
\usepackage{graphicx}
\usepackage{tabulary}
\usepackage{booktabs} % for professional tables

\usepackage{array}
\usepackage{tabularx}
\usepackage[normalem]{ulem}
\usepackage{lipsum}

\usepackage{algpseudocode}
\usepackage{amsmath}
\usepackage{booktabs}
\usepackage{stfloats}
\usepackage{booktabs}

\usepackage[ruled, vlined]{algorithm2e}
\usepackage{tikz}
\usetikzlibrary{decorations.pathreplacing}
\usepackage{arydshln}
\usepackage{enumitem}
\usepackage{algpseudocode}
\usepackage{multirow}
\usepackage{epsfig}
\usepackage{multicol,lipsum,xparse}
\usepackage{caption}
\usepackage{subcaption}
\usepackage{amsmath}
\usepackage{stmaryrd}
\usepackage{color}
\usepackage{todonotes}
\usepackage[utf8]{inputenc}

\copyrightyear{2022}
\acmYear{2022}
\setcopyright{rightsretained}
\acmConference[ICCAD '22]{IEEE/ACM International Conference on Computer-Aided Design}{October 30-November 3, 2022}{San Diego, CA, USA}
\acmBooktitle{IEEE/ACM International Conference on Computer-Aided Design (ICCAD '22), October 30-November 3, 2022, San Diego, CA, USA} \acmDOI{10.1145/3508352.3549378} \acmISBN{978-1-4503-9217-4/22/10}

\begin{document}

%\setcopyright{none}
%\settopmatter{printacmref=false}
\title{Physics-aware Differentiable Discrete Codesign for Diffractive Optical Neural Networks }

\newcommand{\fixme}[1]{\textcolor{red}{\small [~#1~]}}
\author{Yingjie Li, Ruiyang Chen, Weilu Gao, Cunxi Yu}
\email{yingjie.li, cunxi.yu@utah.edu}
\affiliation{%
    \institution{ECE Department, University of Utah, Salt Lake City, Utah, USA}
    \city{Salt Lake City}
    \state{Utah}
    \country{USA}
    \postcode{84112}
    }
 
\settopmatter{printacmref=false}
\begin{abstract}
  Diffractive optical neural networks (DONNs) have attracted lots of attention as they bring significant advantages in terms of power efficiency, parallelism, and computational speed compared with conventional deep neural networks (DNNs), which have intrinsic limitations when implemented on digital platforms. However, inversely mapping algorithm-trained physical model parameters onto real-world optical devices with discrete values is a non-trivial task as existing optical devices have non-unified discrete levels and non-monotonic properties. This work proposes a novel device-to-system hardware-software codesign framework, which enables efficient physics-aware training of DONNs w.r.t arbitrary experimental measured optical devices across layers. Specifically, Gumbel-Softmax is employed to enable differentiable discrete mapping from real-world device parameters into the forward function of DONNs, where the physical parameters in DONNs can be trained by simply minimizing the loss function of the ML task. The results have demonstrated that our proposed framework offers significant advantages over conventional quantization-based methods, especially with low-precision optical devices. Finally, the proposed algorithm is fully verified with physical experimental optical systems in low-precision settings. 
\end{abstract}


\maketitle

\section{Introduction}

During the past half-decade, there has been significant growth in machine learning with deep neural networks (DNNs). DNNs improve productivity in many domains such as large-scale computer vision, natural language processing, and data mining tasks \cite{lecun2015deep,silver2017mastering,senior2020improved}. However, conventional DNNs implemented on digital platforms have intrinsic limitations in computation and memory requirements \cite{jouppi2017datacenter,sharma2016dnnweaver,abadi2016tensorflow}. Moreover, when it deals with computation-intense tasks, its energy cost will be a great concern. To overcome limitations in resources and find an energy-saving computation method, people have turned their eyes to optics \cite{ying2020electronic,lin2018all,zhou2021large,li2020multi,gu2020flops,gu2020towards,gu2021efficient,li2021has,gu2021squeezelight}. Specifically, the free-space diffractive optical neural networks (DONNs), which is based on light diffraction, featuring millions of neurons in each layer interconnected with neurons in neighboring layers, show its great potential in improving efficiency in computing with neural networks \cite{lin2018all}. More importantly, \citet{lin2018all, li2021late, rahman2020ensemble, li2020multi,shen2017deep,chen2022complex} demonstrated that diffractive models controlled by physical parameters are differentiable, such that the parameters can be optimized with conventional automatic differentiation engines.

However, when the DONN system is deployed on physical hardware, it shows significant accuracy degradation compared to the numerical physics emulation \cite{lin2018all,zhou2021large}, e.g., the accuracy degradation is claimed as 30\% in \cite{zhou2021large}. To narrow the algorithm-hardware miscorrelation gaps between differentiable numerical physics models and physical systems, hardware-software codesign training algorithms are needed to deal with the practical response of optical devices. {For example, the reconfigurability of DONNs is implemented using spatial light modulators (SLMs), which have a discrete and non-monotonic complex-valued modulation of propagating optical fields as a function of applied voltages with finite-precision \cite{zhou2021large}.} 

Therefore, despite the diffraction propagation in the DONN system is differentiable, directly adding discrete mapping from device to DONN systems will break the gradient chain in backpropagation. Moreover, in optical hardware systems, diffractive layers implemented with analog optical devices can behave differently due to different optical configurations or device responses, i.e., non-uniformity exists across the compute units (devices), while the DONN model is trained and optimized on digital platforms with uniform and stable number represented computation. Thus, to narrow the gap between numerical emulation and practical deployment, while training a multi-layer DONNs system, there is a great need to develop a flexible training framework that can optimize the DONNs parameters w.r.t various optical devices from layer to layer.
While quantization techniques are applicable to discrete mapping from the device level to DONNs system, there are several critical limitations due to the fact that optical devices used in DONNs are analog, non-monotonic, and non-unified. Specifically, the trainable parameters in DONN systems are not only limited to discrete mappings with irregular and analog device responses but more importantly limited to the constraints in physics. For example, the phase for the light wave is a periodic function with $2\pi$ as the period. Thus, the trainable parameters w.r.t the phase modulation devices in the DONN system should be restricted within $[0, 2\pi]$ and aware of the $2\pi$ period during the training process \cite{lin2018all,zhou2021large,hazan2021ti}.

{This work studies an efficient and flexible codesign framework that enables physics-aware differentiable discrete mappings from devices to DONN systems via Gumbel-Softmax (GS) \cite{maddison2016concrete,jang2016categorical,gumbel1954statistical}, which has been recently experimentally verified on a visible DONNs hardware platform \cite{chen2022physics}. This approach can overcome the aforementioned limitations with GS enabling fully differentiable discrete mapping regardless of the number representation format, range, or discrete distribution. Specifically, our results demonstrate the advantages over existing state-of-the-art DONNs training approaches and quantization algorithms using real-world optical devices, in various DONNs architecture settings. Moreover, we perform comprehensive temperature scheduling exploration and statistical analysis in the GS algorithm to offer insights of this framework. Our results demonstrate the substantial advantages for DONNs co-design in image classification, particularly when deployed optical devices are limited to low precision. Finally, we verify the proposed approach in our visible range DONNs hardware platform \cite{chen2022physics} in low precision settings.}
%\vspace{-5mm}
\section{Background}
\label{sec:background}

\noindent\textbf{Diffractive Optical Neural Networks (DONNs)} -- {Recently, there have been increasing efforts on optical neural networks and {optics-based DNNs hardware accelerators}, which bring significant advantages for machine learning systems in terms of their power efficiency, parallelism and computational speed, demonstrated at various optical computing systems \cite{gao2021graphene,gao2021artificial,mengu2020scale,lin2018all,feldmann2019all,shen2017deep,tait2017neuromorphic,rahman2020ensemble,li2020multi}. Among them, free-space \textit{diffractive optical neural networks} (DONNs), which is based on the light diffraction, features millions of neurons in each layer interconnected with neurons in neighboring layers. The ultrahigh density and parallelism make this system possess fast and high throughput computing capability. One of the significant advantages of DONNs is the computational density and energy efficiency, where such a platform can be scaled up to millions of artificial neurons while with much less energy cost compared to conventional DNNs on digital platforms \cite{lin2018all,li2020multi,mengu2019analysis,mengu2020scale}. 
%In contrast, the design complexity for deploying deep learning algorithms on other optical architectures, e.g., {silicon photonic platform} proposed by \citet{feldmann2019all,feldmann2020parallel,tait2017neuromorphic}, can dramatically increase. 

In conventional DNNs, forward propagations are computed by generating the feature representation with floating-point weights associated with each neural layer. While in DONNs, such floating-point weights are encoded in {the \textbf{complex-valued transmission coefficient of each neuron} in diffractive layers and \textbf{free-space propagation function}} \cite{ersoy2006diffraction}, which is multiplied onto the light wavefunction as it propagates through the neuron {to next diffractive layer}. 
{Specifically, in the numerical emulation for DONN systems, there are two sets of network parameters. One is for diffraction approximation, which is {non-trainable parameters} describing the propagation after the light wave is diffracted at the diffractive layers, and {defined by the natural physics phenomenon.} The diffraction propagation connects neurons between layers. The other one is for phase modulation, which is our targeted ``trainable parameters`` in DONNs. By applying phase modulation to the input light wave at each diffractive layer, the distribution of the light intensity at the end of the system will be modified accordingly. }

Similar to conventional DNNs, the final output class of DONN systems is predicted based on generating labels according to a given one-hot representation, i.e., the max energy reading over the output light intensity of the last layer observed by detectors, where the loss function is at the same time calculated. Thus, the DONN system can be trained by optimizing parameters for phase modulation in each diffractive layer w.r.t specific machine learning (ML) task loss functions, e.g., image classifications.} Details of DONNs training and inference are provided in {Section \ref{sec:result_comp}}.

%However, there are several critical limitations in the existing DONNs training methodology. First, the prediction accuracy of DONNs with few diffractive layers ($\leq 3$) \cite{lin2018all} is highly limited, even with simple image classification datasets. While \citet{lin2018all} claimed that it is caused by fundamental limitations of optical physics, it is believed that there are potentials on improving the complex-domain training algorithms.
{However, when such DONNs systems are deployed on physical hardware, existing training approaches do not take real device response (see Figure \ref{fig:GS_curve}) into consideration and only assume simple phase-only modulation without any limitations. Ideal assumption and quantization errors create miscorrelation gaps between numerical models and hardware deployment, leading to significant accuracy degradation such as 30\% drop on MNIST dataset \cite{zhou2021large}.} %The framework proposed in this work aims to overcome all these limitations with Gumbel-Softmax enabling discrete mapping training.

%Moreover, when such D2NNs system is deployed on physical hardware, existing methodologies cannot deal with the following problems: (1) the accuracy performance shows significant degradation compared with that in emulation (\cite{}), i.e., there is a huge miscorrelation gap between numerical physics models and hardware deployment; (2) the stable optical devices used for phase modulation are mostly fabricated with very limited and non-uniform discrete control states such SLM, Phase Change Material (\cite{}); (3) in a multi-layer D2NNs system where diffractive layers are implemented with different single optical device, even the same devices can perform differently from each other due to the variations and imperfections in fabrications.}

{\noindent\textbf{Physical optical devices for phase modulation} -- The DONN systems will be deployed on physical hardware with analog optical devices after trained on digital platforms. Optical devices functioning as diffractive layers are expected to provide light diffraction and around 2$\pi$ phase modulation over full function range in the DONN system. For example, the Spatial Light Modulators (SLMs)\footnote{\url{https://holoeye.com/lc-2012-spatial-light-modulator}} made with twisted nematic (TN) liquid crystal can provide a phase shift of about $2\pi$ at $450$nm, about $1.8\pi$ at $532$nm and around $1\pi$ at $800$nm w.r.t the full range (e.g., 256 discrete voltage stages) of supplying control voltage. For systems with laser wavelength in Terahertz (THz) range, i.e., $0.1$mm to $1$mm, a 3D printed mask with designed thickness at each pixel made with UV-curable resin can be used as the diffractive layers in THz optical systems \cite{lin2018all}.
%Thus, such SLMs can only function as diffractive layers in the DONN system with laser wavelength in visible range. 

For example, in our DONN system \cite{chen2022physics}, the diffractive layers are implemented with SLMs. Specifically, the SLM is an array of twisted nematic (TN) liquid crystals, which can be twisted to different angles by different applied control voltages, providing different phase modulation for the input light beam. Each pixel in the SLM is a liquid crystal independently controlled by the control voltage, which can be customized by users via HDMI. A practical optical response in amplitude and phase modulation of the SLM with eight discrete control voltages is provided in Figure \ref{fig:GS_curve}, and we can see that the optical responses w.r.t discrete control voltages is not unified over its functioning range. Additionally, each single SLM can response differently even under identical experimental setups due to fabrication errors. When multiple SLMs are employed in one system, the error will accumulate, and worsening the correlation between the numerical emulations and the physical hardware experiments, which highlights the importance and motivation of our proposed hardware-software codesign framework.}

%of the precise computation kernels for emulation and the hardware-software codesign algorithms designed specifically for DONN systems.}

\noindent\textbf{Gumbel-Softmax} -- {Gumbel-Softmax is a continuous distribution on the simplex which can be used to approximate discrete samples \cite{maddison2016concrete,jang2016categorical,gumbel1954statistical}. With Gumbel-Softmax, discrete samples can be differentiable and their parameter gradients can be easily computed with standard backpropagation. 
Let $z$ be the discrete sample with one-hot representation with $k$ dimensions and its class probabilities are defined as $\pi_1, \pi_2, ..., \pi_k$. Then, according to the Gumbel-Max trick proposed by \citet{gumbel1954statistical}, the discrete sample $z$ can be presented by:
\begin{equation}
\small
    z = \texttt{one\_hot}(\underset{i}{\text{argmax}}[g_i + log{\pi_i}])
    \label{equ:z}
\end{equation}
where $g_i$ are i.i.d samples drawn from Gumbel(0, 1). Then, we can use the differentiable approximation \texttt{Softmax} to approximate the one-hot representation for $z$, i.e., $\nabla_{\pi} z \approx \nabla_{\pi} y$:
\begin{equation}
\small
    y_i = \frac{\text{exp}((\text{log}(\pi_i) + g_i)/\tau)}{\sum_{i=1}^{k} \text{exp}((\text{log}(\pi_i) + g_i)/\tau)} 
\end{equation}

where $i = 1, 2, ..., k$. The softmax temperature $\tau$ is introduced to modify the distributions over discrete levels. Softmax distributions will become more discrete and identical to one-hot encoded discrete distribution as $\tau \rightarrow 0$, while at higher temperatures, the distribution becomes more uniform as $\tau \rightarrow \infty$ \cite{jang2016categorical}. Gumbel-Softmax distributions have a well-defined gradient $\frac{\partial{y}}{\partial{\pi}}$ w.r.t the class probability $\pi$. When we replace discrete levels with Gumbel-Softmax distribution depending on its class probability, we are able to use backpropagation to compute gradients. Recently, Gumbel-softmax has been applied to differentiable neural architecture search \cite{wu2019fbnet,wu2018mixed,he2020milenas,fu2021auto,fu2021a3c} and differentiable quantization \cite{baevski2019vq}. However, enabling differentiable discrete weight training in a fully physics-differentiable neural networks with real-world physical system has not yet been studied. Particularly, the parameters in DONNs are limited to non-negative values due the nature of optical physics. 
}

%\paragraph{Hardware Device} {In D2NN system, we can realize the phase modulation by employing phase-change devices, such as Spatial Light Modulator (SLM), Phase Change Material (PCM) and 3D printed phase mask. These devices have limited and discrete modulation stages due to the intrinsic physical limitations. The phase modulation provided by SLM is controlled by the applied voltage to each pixel and due to its imperfections, the SLM will also modify the intensity of the input beam of light. The voltage applied can be customized by entering the level which is a integer number set in the corresponding software. Thus, SLM can only modulate the input light beam with discrete states, e.g., for a specific SLM, the phase modulation ranging $2\pi$ is divided into 256 levels, so that users can customize the modulation by entering the integer number between 0 and 255, i.e., the SLM only works with discrete voltage levels. Meanwhile, due to the fabrication variations, SLMs even with the same manual can have different function curves. Thus, we need to calibrate the device manually to generate accurate data points for system training. For 3D printed phase masks, the modulation precision is determined by the fabrication precision of the 3D printer and the intensity modulation is determined by the material used for fabrication. For PCM, its stable functionality for phase modulation can only be realized at few levels, which means it has significantly limited discrete states for hardware deployment. }
\section{Approach}
\label{sec:approach}

\begin{figure*}[t]
     \centering
     \begin{subfigure}[b]{0.9\textwidth}. 
         \centering
         \includegraphics[width=0.8\linewidth]{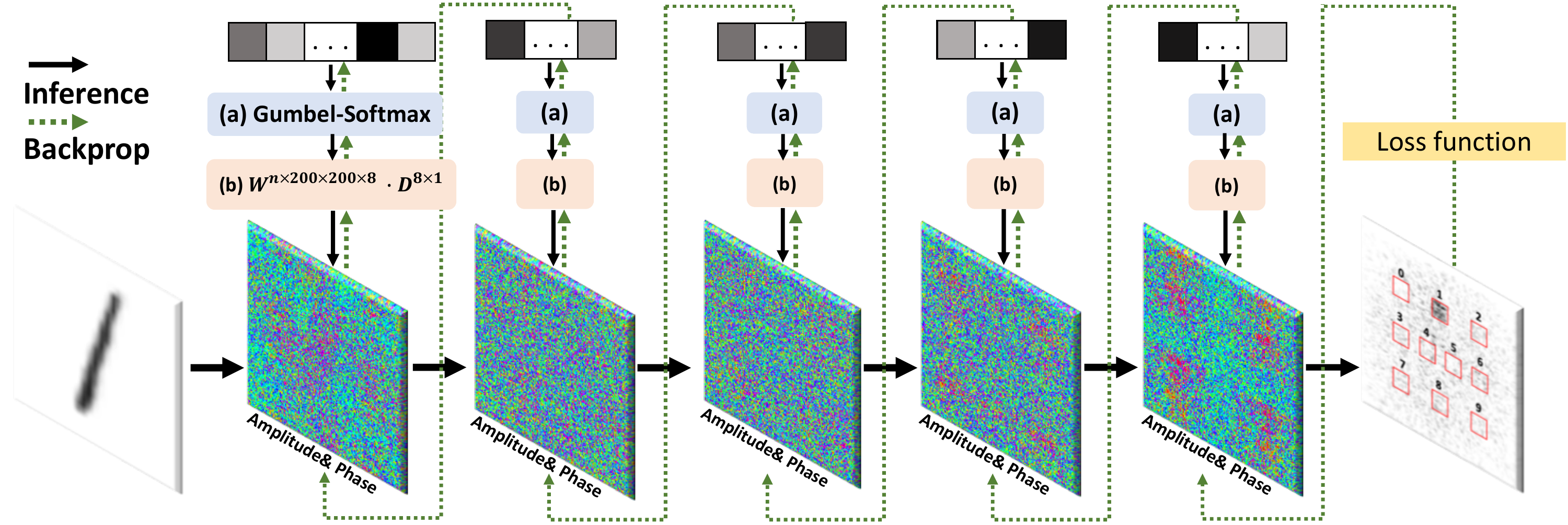}
         \caption{}
         \label{fig:sys_setup}
     \end{subfigure}
     \hfill
     \begin{subfigure}[b]{0.75\textwidth}
         \centering
         \includegraphics[width=0.8\linewidth]{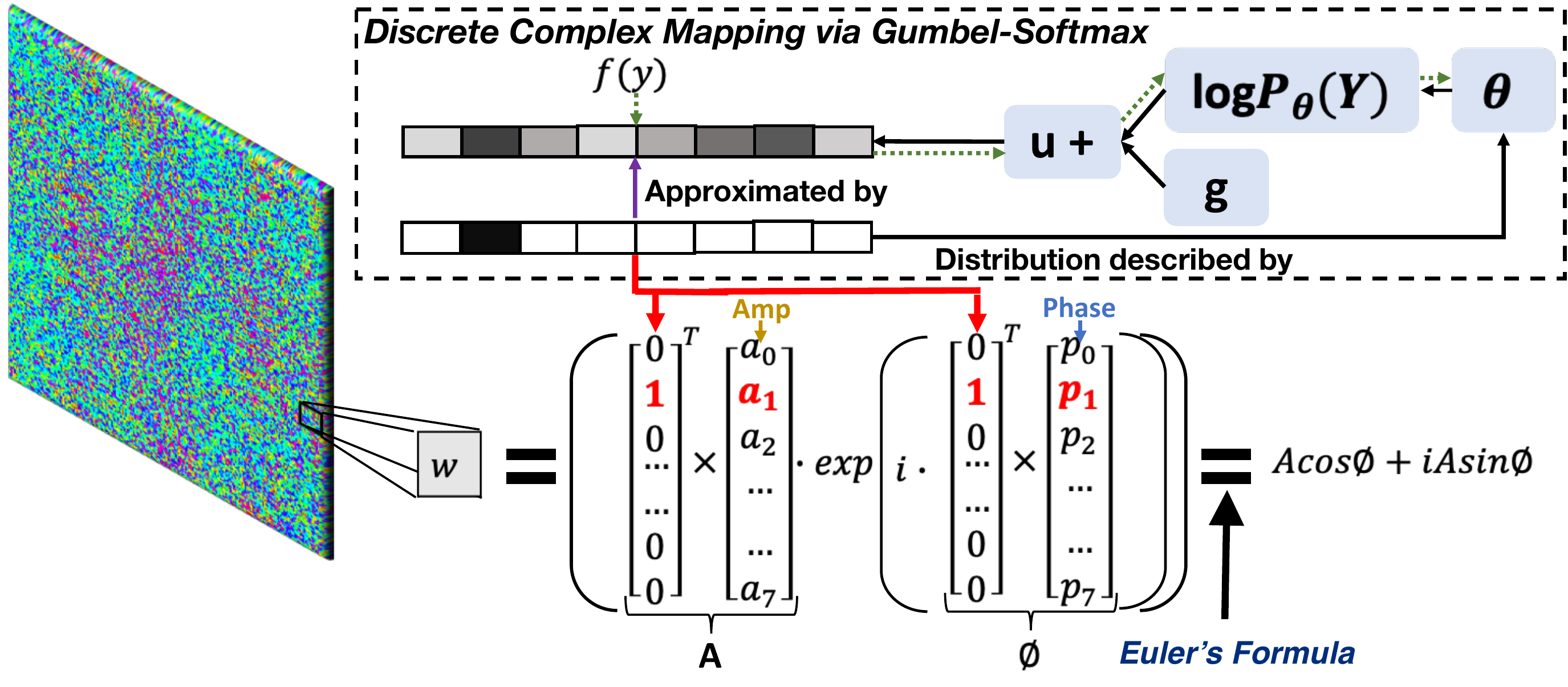}
         \caption{}
         \label{fig:sys_detail}
     \end{subfigure}
     \hfill
     \begin{subfigure}[b]{0.24\textwidth}
         \centering
         \includegraphics[width=0.9\linewidth]{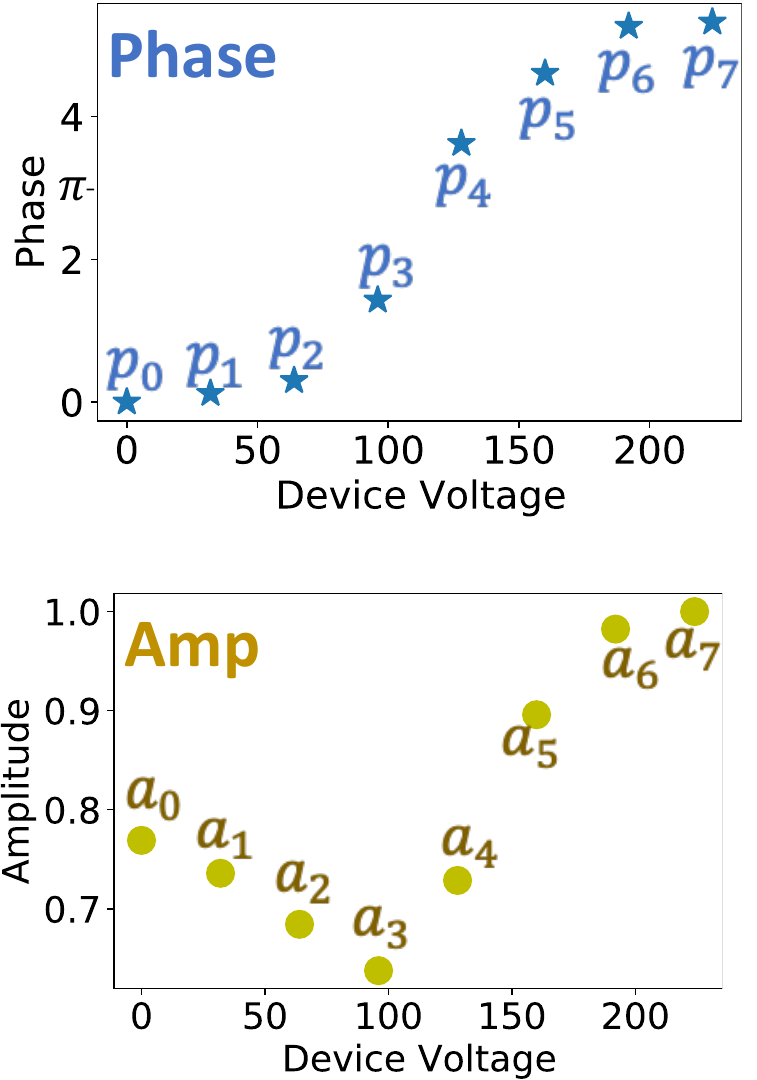}
         \caption{}
         \label{fig:GS_curve}
     \end{subfigure}
        \caption{{Illustration of the proposed GS framework in training a 5-layer DONN system} -- (a) During inference, input signal is modulated by each layer in both amplitude and phase encoded by optical devices controlled by discrete voltage values. In backpropagation for training, the gradients will flow through Gumbel-Softmax over the approximated differentiable representation of the one-hot encoded discrete voltage levels; (b) Detailed illustration of discrete device-to-system mapping via Gumbel-Softmax in both forward and backpropagation; (c) Illustration of real-world device responses.}
        \label{fig:system}
\end{figure*}
 
%This section describes the proposed training framework integrated with Gumbel-Softmax, which enables fully differentiable physic-aware training with real-world optical devices.

\textbf{System Overview and GS-based Training Algorithm} --
%The DONNs system implemented in this work is composed of five diffractive layers and one detector with ten detector regions, representing ten classes for classification, placed evenly on the detector plane. The optical devices used for deploying phase and amplitude modulation are controlled by the input discrete voltage values. As shown in Figure \ref{fig:system}, each diffractive layer modifies the amplitude and phase of the input light signal. First, our framework initializes the input discrete voltage levels for each pixel in layers as an one-hot representation. The one-hot vector size will be identical to the number of data points calibrated manually from the device, e.g., when a system is implemented with an optical device in a system whose diffractive layers have the size of $200 \times 200$ and  the employed device has the discrete voltage levels of $8$.  we will have are two $8 \times 1$ arrays for hardware-reference data by calibrating the device manually, one for phase modulation and the other one for amplitude modulation. Then the layer size will be one-hot encoded with the size of $200 \times 200 \times 8$ by one-hot encoding each pixel in the layer with size of $8$. The phase and amplitude modulation information in the device will be generated with the size of $200 \times 200 \times 1$ by \texttt{matmul} the one-hot represented layer and the calibrated data points.
We illustrate the proposed training framework using a five-layer DONN system implemented for a ten-class image classification task, with ten detector regions placed evenly on the detector plane. As shown in Figure \ref{fig:system}, each diffractive layer modifies the amplitude and phase of the input light signal. In our experimental setup, diffractive layers are implemented with SLMs. As we discussed in Section \ref{sec:background}, the analog optical device SLMs provide modulation w.r.t the discrete input control voltage values, and the physical responses for each SLM can be different and non-unified, which requires a hardware-software codesign algorithm for precise emulation. 

To enable differentiable discrete mapping, \textit{Gumbel-Softmax} is added in the numerical modeling of DONNs. First, our framework defines the input discrete voltage values as trainable parameters, where each pixel is represented using a one-hot vector. The trainable parameters dimensions are then defined by (1) the system size and (2) the number of discrete values in the devices. For example, let the system size be $200 \times 200$ with $8$ discrete levels in the devices, the trainable voltage parameters will be $200 \times 200 \times 8$ in each layer. Note that the phase and amplitude modulation will still be in the shape of $200 \times 200$, where the optical properties are mapped by \texttt{matmul} the one-hot vectors ($200 \times 200 \times 8$) and the device level vector ($8 \times 1$). Then, to deal with the problem of gradient chain breakage brought by the discrete trainable parameters, in Gumbel-Softmax, a Gumbel distribution $g \sim Gumbel(0, 1)$ and a class probability $\theta$ for the discrete levels are introduced to approximate the discrete levels. As it is shown in Figure \ref{fig:sys_detail}, during backpropagation in the training process, instead of propagating gradients to the discrete one-hot voltage levels directly, it will propagate through the differentiable approximation to the discrete levels generated by Gumbel-Softmax distribution with its class probability $\theta$. The differentiable approximation will be updated according to the training algorithm, and the discrete voltage levels will be updated by its class probability $\theta$ from approximation. The process in Figure \ref{fig:sys_detail} can be described as:
\begin{equation} \label{eq:gumbel-voltage}
\begin{split}
\small
&w^{i,j} = \texttt{one\_hot}( \frac{\text{exp}((\text{log}(\theta_{n}) + g_{n})/\tau)}{\sum_{n=1}^{k} \text{exp}((\text{log}(\theta_{n}) + g_{n})/\tau)}), \\ &g_{n} \sim \text{Gumbel}(0,1), ~k = \text{total discrete levels}, ~i,j \in [0, N-1] \\
\end{split}
\end{equation}
where $w^{i, j}$ be the voltage level applied to the pixel located at $[i, j]$ in the diffractive layer with a size $N \times N$. In Equation \ref{eq:gumbel-voltage}, \texttt{one\_hot} will pick the class with the highest probability over $k$ discrete voltage levels after softmax as $1$, while other $k-1$ stages will be $0$ in the one-hot representation for $w^{i, j}$.
\begin{equation} \label{eq:gumbel-phase}
\begin{split}
\small
w_{\mathbb{C}}^{i,j} &=  \overbrace{\underbrace{(w^{i,j} \cdot A)}_{\text{Matmul}} \cdot \text{cos}(w^{i,j} \cdot P)}^{\text{Real}} + \underbrace{i(w^{i,j} \cdot A) \cdot \text{sin}(w^{i,j} \cdot P)}_{\text{Imaginary}}, \\ & \text{for}~i,j \in [0, N-1]% \\
%\Re(z) and \operatorname{Re}(z)
\end{split}
\end{equation}
Let $A$ be the array with discrete calibrated amplitude value, $P$ be the array with discrete calibrated phase value. $w_{\mathbb{C}}^{i,j}$ will be the phase and amplitude modulation provided by the pixel located at $[i, j]$, which is a complex number resulting from mapping the one hot represented voltage level with the amplitude and phase responses of the specific SLM (e.g., the optical responses in Figure \ref{fig:GS_curve}). Let $w^{i,j} \in W$, $w^{i,j}_{\mathbb{C}} \in W_{\mathbb{C}}$, where $i,j \in [0, N-1]$, $N$ is the size of diffractive layer. According to Equation \ref{eq:gumbel-voltage}, the discrete variable $W$ has the distribution depending on $\theta$ and forward function $f(W)$. The objective is to minimize the expected cost $L(\theta) = {\mathbb{E}}_{W \sim p_{\theta}(W)}[f(W)]$, which is the ML loss in the system, e.g., in DONNs system for image classification, $L$ is usually set as the MSE Loss \cite{lin2018all}, \cite{zhou2021large}, via gradient descent, which requires us to estimate $\nabla_{\theta}\mathbb{E}_{W \sim p_{\theta}(W)} [f(W)]$. The discrete sample $W$ can be approximated by $G(\theta, g)$. The gradients from $f$ to $\theta$ will be computed as follows:
\begin{equation} \label{eq:optimization}
\begin{split}
\frac{\partial}{\partial{\theta}}\mathbb{E}_{W\sim p_{\theta}(W)}[f(W)] = \frac{\partial}{\partial {\theta}}\mathbb{E}_{g}[f(G(\theta, g))] \\ = \mathbb{E}_{g \sim \text{Gumbel}(0, 1)}[\frac{\partial f}{\partial G} \frac{\partial G}{\partial \theta}]
\end{split}
\end{equation}
To be more specific, as shown in Figure \ref{fig:sys_detail}, for each pixel ($w_{\mathbb{C}}^{i, j}$) in diffractive layers, the modulation provided by the pixel is represented by complex tensors that emulate the diffraction of light. The complex number is transformed from phase and amplitude modulation by {\it Euler's formula}. Since the input light signal is also described by complex numbers, the modulation can be easily realized by multiply the two complex numbers. 

As a results, after the input image is encoded with a coherent laser beam in complex domain, where the imaginary part is initialized as all zeros, the emulations/forward function of the DONN system for image classification can be described as follows -- Each diffractive layer is composed with dense diffraction units with apertures and phase modulators. Specifically, the aperture will provide \textbf{light diffraction}, which connects the neurons in neighboring diffractive layers and the diffraction is non-trainable parameters w.r.t user-defined setups in the system. The \textbf{phase modulation} embedded in each aperture functions as 'neurons' in DONN to modify the input light wave, and the phase modulation is our targeted trainable parameters in the system.
In the emulation for the physics-based DONN system, we employ \textbf{(1)} Fresnel approximation \cite{ersoy2006diffraction}, which is a mathematical approximation for light diffraction through an aperture, in the forward function to describe the light diffraction in DONN systems; \textbf{(2)} complex-valued matrix multiplication for modulator parameters and the input light wavefunction to describe the phase modulation in the system. 

For light diffraction, the input at point $(x, y)$ at $l-$th diffractive layer can be seen as the summation of the outputs at $(l-1)-$th layer over the plane $(x', y')$, i.e.,
\begin{equation}
\label{eq:diffraction_time}
    f_{l}^{1}(x, y, z) = \iint f_{l-1}^{0}(x', y', 0)h(x-x', y-y', z)dx'dy'
\end{equation}
where $z$ is the distance between layers, $h$ is the impulse response function of free space. $f_{l-1}^{0}$ is the output wavefunction of points on $(l-1)-$th layer and also the input for free-space propagation, $f_{l}^{1}$ is the output function of the free-space propagation and the input to the phase modulation at $l-$th layer. Equation \ref{eq:diffraction_time} can be calculated with spectral algorithm, where we employ Fast Fourier Transform (FFT) for fast and differentiable computation. By convolution theorem, the integral over $x$, $y$-axes of the convolution of $f_{l-1}^{0}$ and $h$ is the product of 2D Fourier transformations over $x$, $y$-axes of $f_{l-1}^{0}$ and $h$, i.e., 
\begin{equation}
    \mathcal{F}_{xy}(f_{l}^{1}(x, y, z)) =  \mathcal{F}_{xy}(f_{l-1}^{0}(x', y', 0))\mathcal{F}_{xy}(h(x, y, z))
\end{equation}
\begin{equation}
\label{equ:diffraction_freq}
    U_{l}(\alpha, \beta, z) = U_{l-1}(\alpha, \beta, z)H(\alpha, \beta, z)
\end{equation}
where $U$ and $H$ are the Fourier transformation of $f$ and $h$ respectively. The impulse function used for Fresnel approximation is 
\begin{equation}
\label{eq:diffraction_time_end}
    h(x, y, z) = \frac{\exp(ikz)}{i\lambda z}\exp\{\frac{ik}{2z}(x^{2} + y^{2})\}
\end{equation}
where $i=\sqrt{-1}$, $\lambda$ is the wavelength of the laser source, $k=2\pi/\lambda$ is free-space wavenumber. 

For phase modulation, we model the output of the free-space propagation at $l-$th layer from Equation \ref{equ:diffraction_freq} as $U_{l}(\alpha, \beta, z)$, which is then converted back to spatial domain $(x, y)$ via inverse FFT as $f_{l}^{2}(x, y, z)$ as the input to the phase modulation at $l-$th layer. The phase modulation is described in Equation \ref{eq:gumbel-phase} as $W_{\mathbb{C}}$. As we discussed in Section \ref{sec:background}, the phase modulation provided by each diffraction unit in the layer is independently configured, thus, the phase modulation $W_{\mathbb{C}}$ functions w.r.t the location $(x, y)$ at the $l-$th layer. Thus, the output wavefunction after phase modulation is expressed as
\begin{equation}
\label{eq:phase_mod}
    f_{l+1}^{0}(x, y, 0) = f_{l}^{2}(x, y, z) \times W_{\mathbb{C}}^{l}(x, y)
\end{equation}
where $f_{l+1}^{0}(x, y, 0)$ is the input wavefunction for the forward function for $(l+1)-$th layer.

As a result, we model the light diffraction (Equations \ref{eq:diffraction_time}--\ref{eq:diffraction_time_end}), phase modulation (Equation \ref{eq:phase_mod}), and device-to-system codesign (Equations \ref{eq:gumbel-voltage}--\ref{eq:gumbel-phase}) in fully differentiable numerical formats. Thus, the DONNs can be trained with conventional \textit{autograd} algorithms by {simply minimizing DONNs ML loss function}, e.g., Adam algorithm \cite{kingma2014adam}. Note that in the DONN system, since diffractive layers are propagated in sequence, each layer can be implemented using different devices, i.e., with different calibrated data points. As we can see in Figure \ref{fig:system}, mapping multiple devices in one DONN system can be simply realized using the proposed framework by replacing the $Amp$ and $Phase$ vectors at each layer.

\noindent\textbf{Gumbel-Softmax Exploration} --
As it is discussed in Section \ref{sec:background}, the variance of the approximated Gumbel-Softmax distribution over discrete levels is determined by $\tau$, which is also referred as \textit{temperature} in Gumbel-Softmax. While giving a higher temperature value, Gumbel-Softmax will result in less variance that is close to a uniform distribution. In opposite, when $\tau$ is close to 0, it will be more variant over discrete levels, i.e., be more identical to one-hot distribution. Thus, similar to simulated annealing, the algorithm should be first deployed with high temperature to enable coarse-grain global search. The temperature should then be annealed down to shrink the search space to find the local optimized point. Specifically, at the early training stage, we expect the variance between different levels to be small, such that the discrete values are easier to be changed during gradient descent optimization. As the optimization efforts increase, it is expected to decrease the temperature to fine-tune the optimization, where most of the discrete values are far more stable during gradient descent optimization. In Section \ref{sec:result_temp}, we explore and provide comprehensive discussions on six different temperature schedules.

%\vspace{-4mm}
\section{Results}
\label{sec:results}

\begin{figure*}[!htb]
    \centering
    \begin{subfigure}[b]{0.49\linewidth}
    \centering
        {\includegraphics[width=0.8\linewidth]{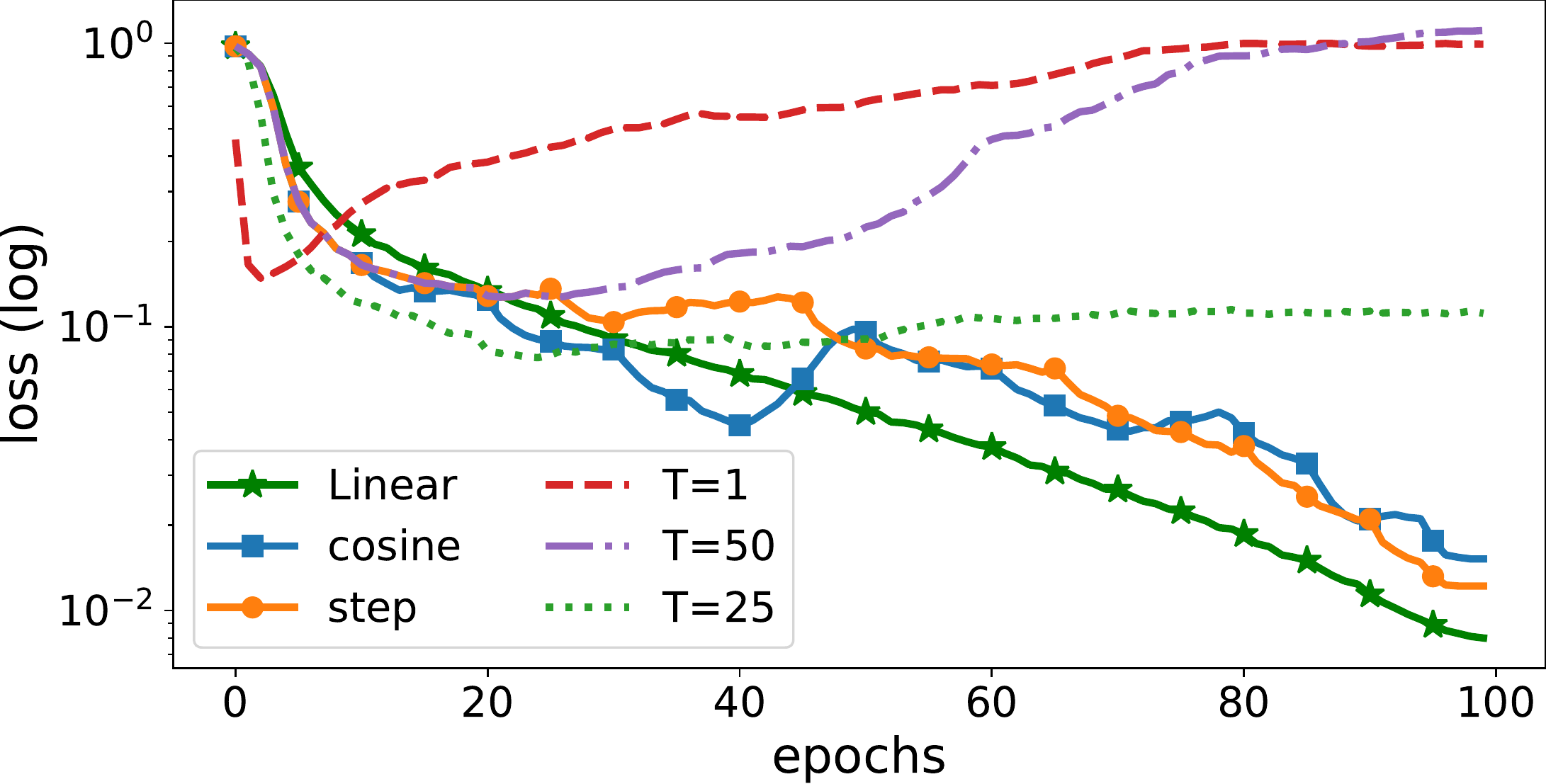}}
        \subcaption{Training loss evaluations on MNIST.}
    \end{subfigure}
    \hfill
    \begin{subfigure}[b]{0.49\linewidth}
    \centering
        {\includegraphics[width=0.8\linewidth]{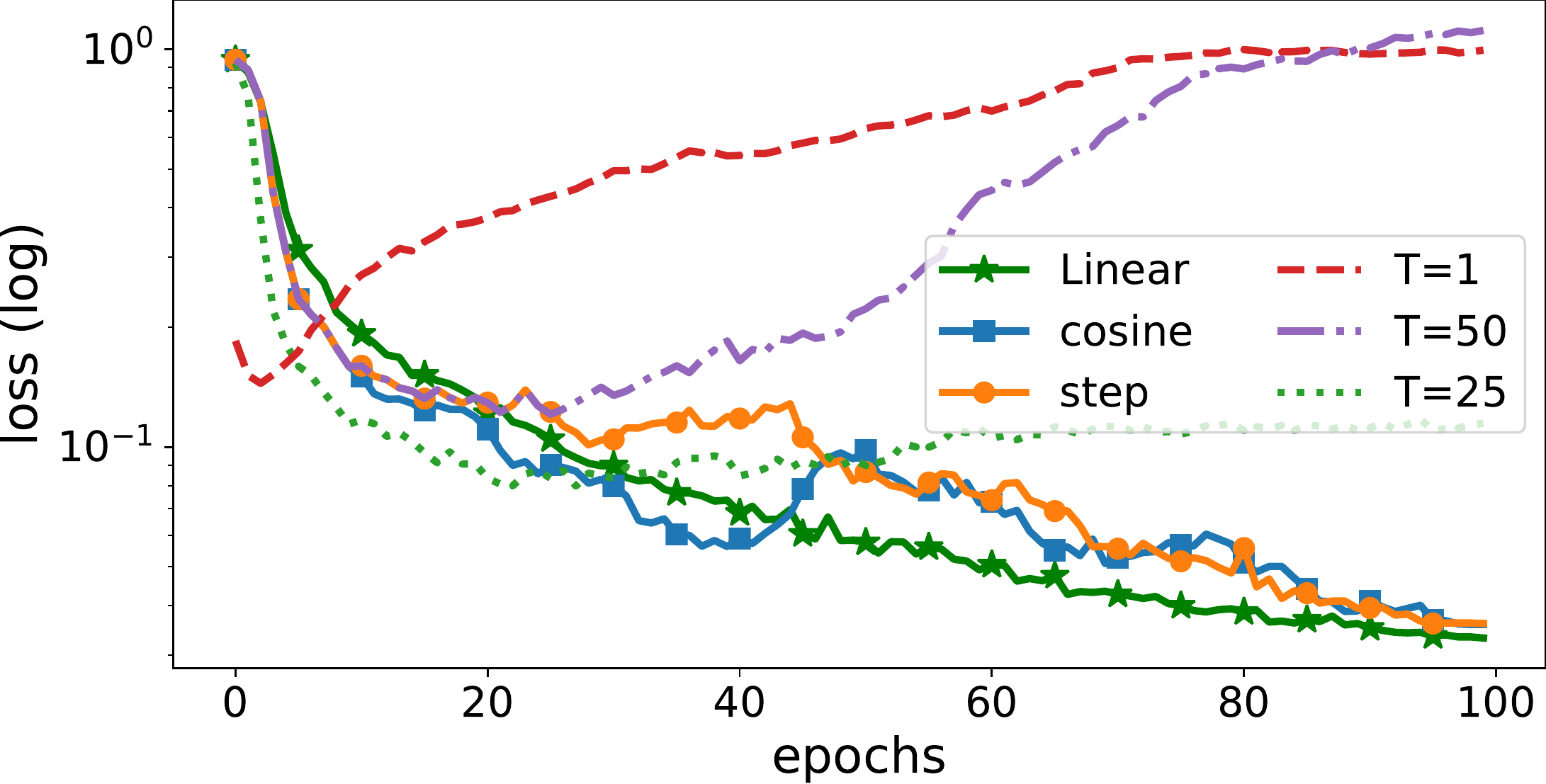}}
        \subcaption{Testing loss evaluations on MNIST.}
    \end{subfigure}
    \hfill
    \begin{subfigure}[b]{0.49\linewidth}
    \centering
        {\includegraphics[width=0.8\linewidth]{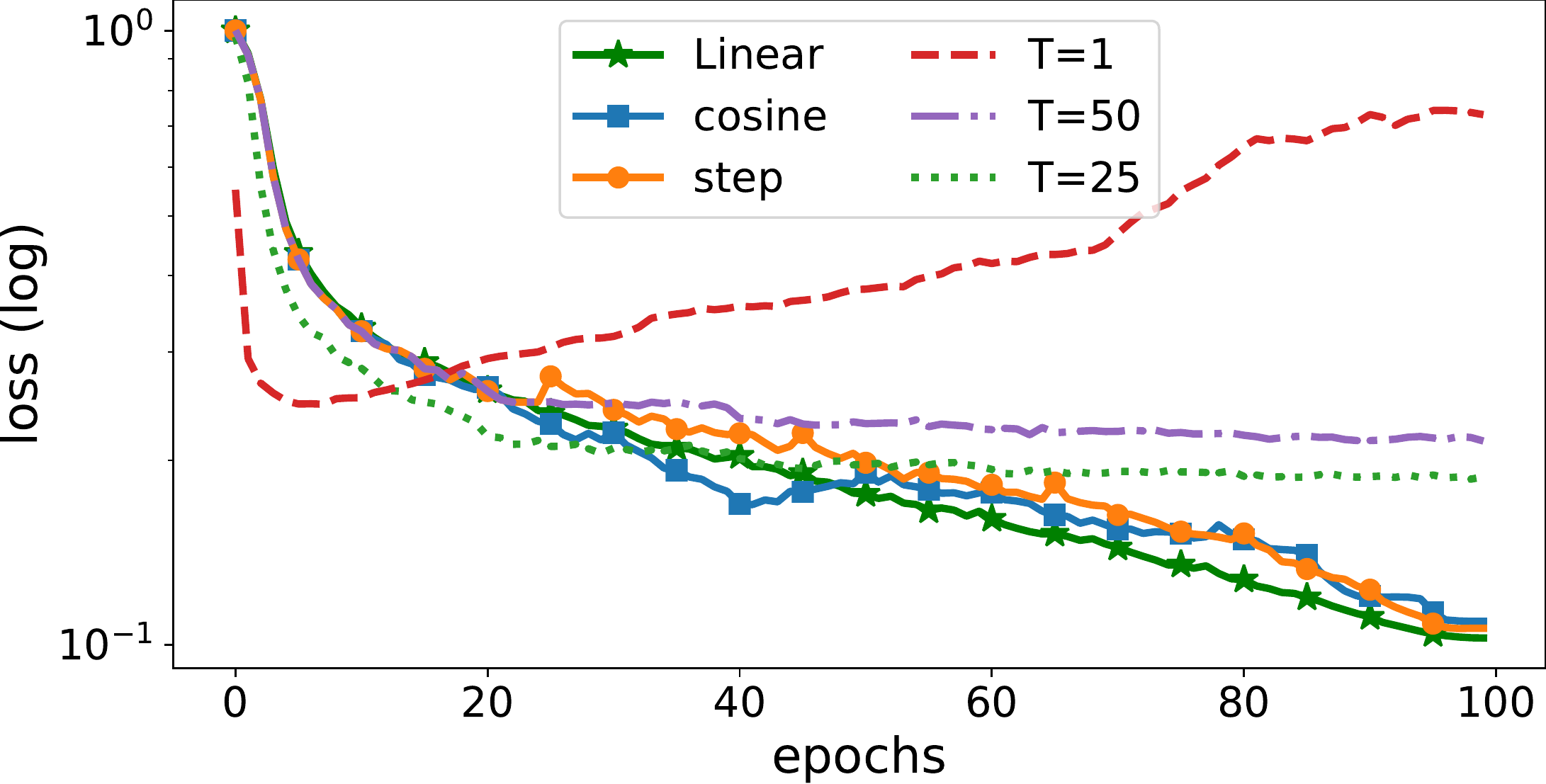}}
        \subcaption{Training loss evaluations on F-MNIST.}
    \end{subfigure}
    \hfill
    \begin{subfigure}[b]{0.49\linewidth}
    \centering
        {\includegraphics[width=0.8\linewidth]{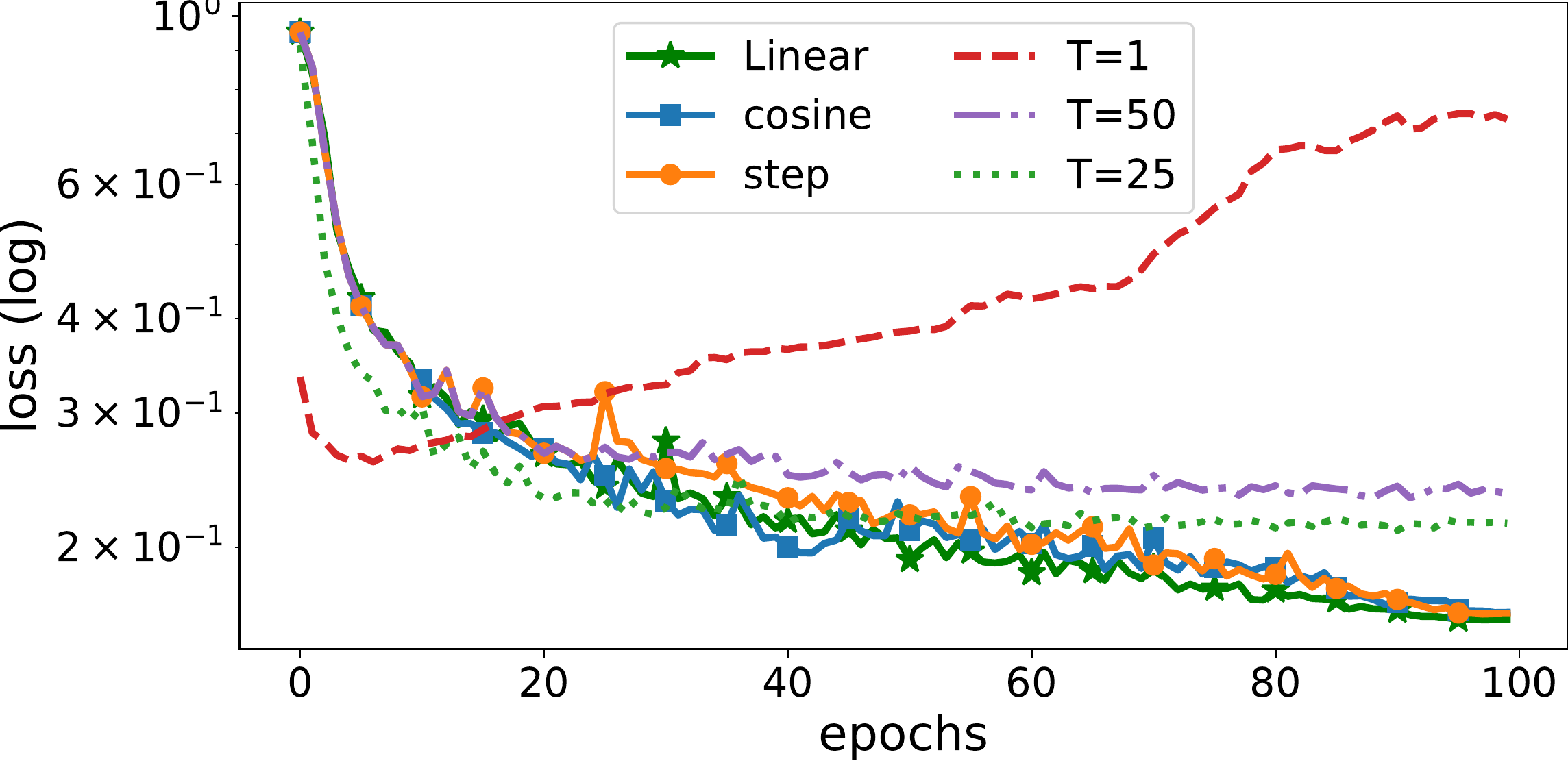}}
        \subcaption{Testing loss evaluations on F-MNIST.}
    \end{subfigure}

\vspace{2mm}
\begin{subfigure}{1\linewidth}
\centering
{
\begin{tabular}{|c|c|c|c|c|c|c|}
    \hline
      & 
     \texttt{Linear} & \texttt{Cosine} & \texttt{Step} & $\tau=50$ & $\tau = 25$ & $\tau = 1$ \\
     \hline
     MNIST & 0.997/0.979 & 0.994/0.978 & 0.995/0.978 & 0.138/0.133 & 0.930/0.926 & 0.143/0.146\\
     \hline
     FMNIST & 0.941/0.891 & 0.935/0.888 & 0.938/0.889 & 0.857/0.844 & 0.876/0.854 & 0.446/0.444\\
     \hline
\end{tabular}
}
		\caption{Final training/testing accuracy of all temperature schedules.}
	\end{subfigure}
    \caption{Evaluations of temperature scheduling in GS-based discrete device-to-system DONNs training framework.}
    \vspace{-3mm}
    \label{fig:temp}
\end{figure*}
\textbf{System Parameters} -- The default system used in this work is designed with five diffractive layers with the size of $200 \times 200$, i.e., the size of layers and the size of total ten detector regions are $200 \times 200$. To fit the optical system, the original input images from MNIST \cite{lecun1998mnist} and FashionMNIST (FMNIST) \cite{xiao2017fashion} with size of $28 \times 28$ will be interpolated into size of $200 \times 200$ and encoded with the laser source whose wavelength is $532~nm$. The physical distances between layers, first layer to source, and final layer to detector, are set to be $27.94~cm$. As shown in Figure \ref{fig:system}, ten separate detector regions for ten classes are placed evenly on the detector plane with the size of $20 \times 20$, where the sums of the intensity of these ten regions are equivalent to a 1$\times$10 vector in \texttt{float32} type. The final prediction results will be generated using \texttt{argmax}. The default DONN system is implemented with an optical device consisting of 8 discrete voltage values for modulation.  \textbf{Training Setups} -- The learning rate in the training process is $0.5$ trained with $100$ epochs for all experiments using \texttt{Adam} \cite{kingma2014adam} with batch size $500$. The implementations are constructed using {PyTorch v1.8.1}. All experimental results are conducted on an Nvidia 2080 Ti GPU.

\begin{figure*}[h]
    \centering
    \begin{subfigure}[b]{1\linewidth}
    \centering
        {\includegraphics[width=0.86\linewidth]{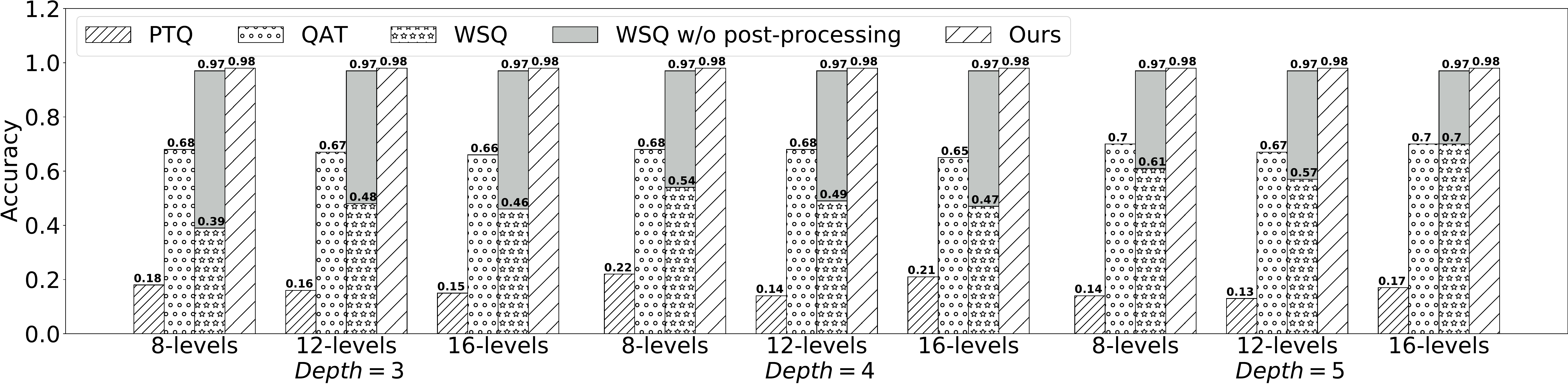}}
        \caption{{Evaluation on} 3-layer, 4-layer, and 5-layer DONNs with MNIST-10.}
    \end{subfigure}
    \hfill
    \begin{subfigure}[b]{1\linewidth}
        \centering
        {\includegraphics[width=0.86\linewidth]{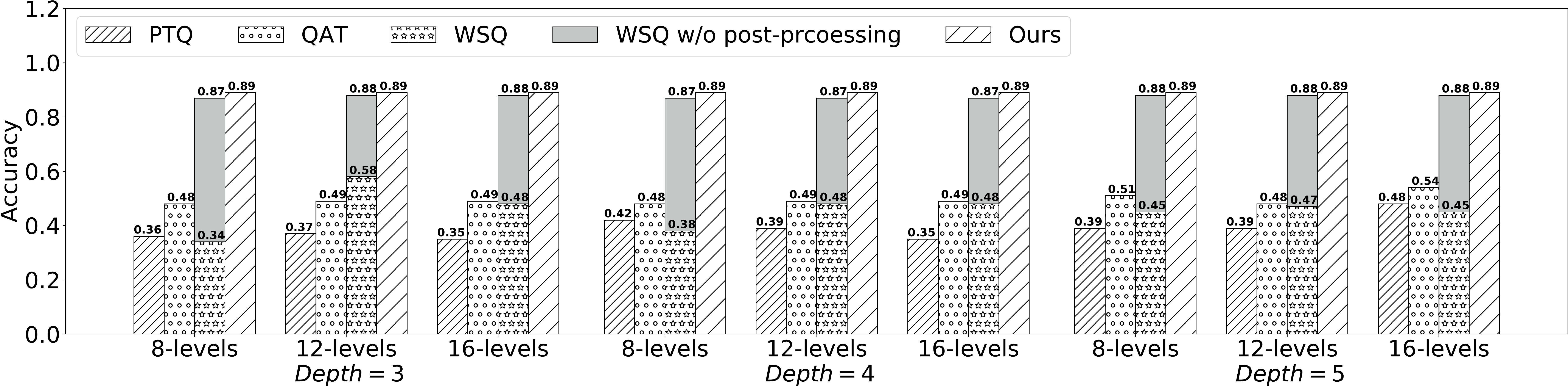}}
        \caption{{Evaluation on} 3-layer, 4-layer, and 5-layer DONNs with FashionMNIST.}
    \end{subfigure}
    \caption{Accuracy performance comparisons between the proposed framework and conventional quantization methods, i.e., PTQ \cite{krishnamoorthi2018quantizing}, QAT \cite{krishnamoorthi2018quantizing}, and WSQ \cite{han2015deep, ullrich2017soft}, for discrete device mapping in 3-, 4-, and 5-layer DONNs using MNIST and FMNIST datasets. WSQ w/o post-processing shows the accuracy performance of the model quantized with both negative and positive discrete levels. Ours denotes the performance of the model quantized with the proposed GS framework.}
    \vspace{-3mm}
    \label{fig:quan_vs_gumble}
\end{figure*}

\vspace{-4mm}
\subsection{Temperature scheduling in Gumbel-Softmax}
\label{sec:result_temp}

While deploying Gumbel-Softmax to enable differentiable discrete training for DONNs, the temperature value in Gumbel-Softmax is known as an important hyperparameter for the training performance. Thus, we first explore different temperature schedules in the proposed training framework for implementing DONNs with low-precision optical devices. Specifically, the results shown in Figure \ref{fig:temp} are trained with experimentally measured real-world devices with 8 discrete values (see Figure \ref{fig:GS_curve}). 
%{\color{blue} \%}As discussed in Section \ref{sec:approach}, higher temperature leads to less variation between the discrete levels, while lower temperature leads to a distribution closer to categorical distribution. Specifically, with higher temperature implemented in Gumbel-Softmax, a larger range of possible levels will be explored as the distribution is more uniform over all possible levels. 

As discussed in Section \ref{sec:approach}, we mimic the concept of temperature scheduling in \textit{simulated annealing} into the Gumbel-Softmax based training framework. We evaluate six different temperature schedules, which all start with highest temperature $\tau_{h}=50$, and end with lowest temperature $\tau_l=1$ with 100 training epochs (Figure \ref{fig:temp}). First, we set three static temperature training as baselines for comparisons, which are trained with static temperatures for the whole training process, i.e., (1) $\tau=1$, (2) $\tau=25$, and (3) $\tau=50$. For dynamic temperature scheduling, we evaluate (4) linear temperature decaying scheduling (\texttt{Linear}), with temperature decay rate as $0.5$ per epoch; and (5) cosine-annealing-decaying (\texttt{Cosine}) temperature schedule, where we set $\tau_{cosine}=[50, 40, 30, 20, 40, 30, 20, 30, 15, 5, 10, 1]$. With higher temperature, i.e., larger exploration space for the algorithm, it is expected to train more epochs. Thus, we set the training epochs for each temperature as $[10, 10, 10, 10, 10, 10, 10, 8, 7, 5, 5, 5]$ ; and (6) step temperature decaying (\texttt{Step}) the temperature schedule is set as $\tau_{step}=[50, 40, 30, 20, 10, 5, 1]$ with the training epochs per temperature as $[25, 20, 20, 15, 10, 5, 5]$.

In Figures \ref{fig:temp}(a)--(d), the training and testing processes are fully recorded for MNIST and FMNIST, with the x-axis representing the training iteration and the y-axis representing the loss value. The training and testing accuracy is shown in Figure \ref{fig:temp}(e). As we can see, for both MNIST and FMNIST, the system performs better with annealing temperature schedules in Gumbel-Softmax compared to the static temperature setups. From Figure \ref{fig:temp}, we empirically conclude that the \texttt{Linear} annealed temperature schedule works best for our system, as training and testing loss converge simultaneously and most efficiently. Therefore, we deploy \texttt{Linear} temperature decaying in the rest experiments. %Specifically, in Figure \ref{fig:temp}, we can see that: \textbf{(1)} The system implemented with simulated annealing applied Gumbel-Softmax performs better than that implemented with static temperature applied Gumbel-Softmax. With bad static temperature implemented (e.g., $\tau=50$ and $\tau=1$ for MNIST dataset and $\tau=1$ for FMNIST dataset), the system can get worse as training efforts increases. \textbf{(2)} The \texttt{Linear} temperature annealing schedule works best for our system, as training and testing loss converge simultaneously and most efficiently. Therefore, we deploy \texttt{Linear} temperature decaying in the rest of the section.

\begin{figure}[!htb]
    \centering
    \begin{subfigure}[b]{0.48\linewidth}
    \centering
        {\includegraphics[width=1\linewidth]{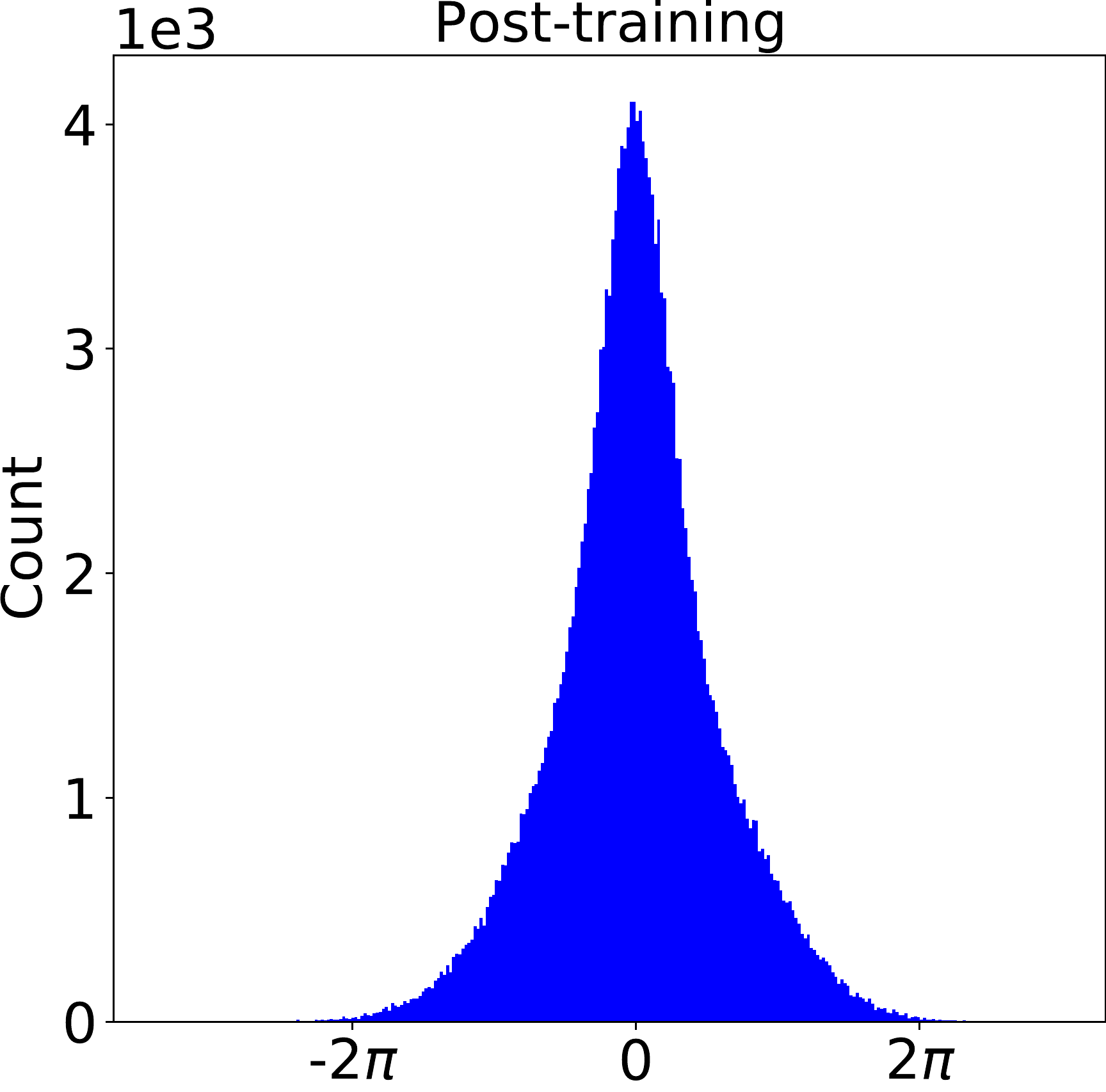}}
        %\caption{Distribution for full precision \texttt{float32}}
         \caption{Full precision (\texttt{float32}).}
         \label{fig:weight_full}
    \end{subfigure}
    \hfill
    \begin{subfigure}[b]{0.48\linewidth}
    \centering
        {\includegraphics[width=1\linewidth]{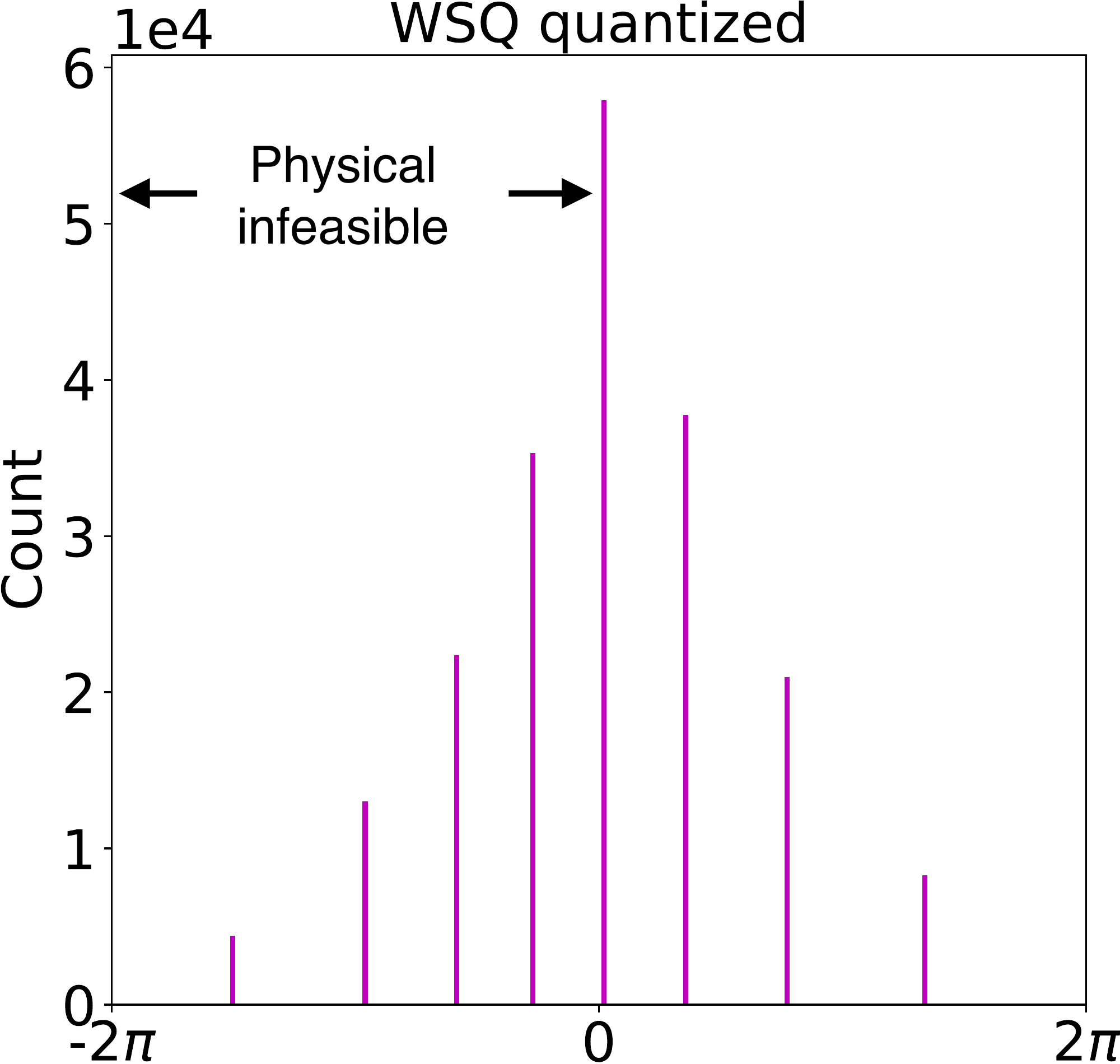}}
        \caption{WSQ w/o post-processing.}
        \label{fig:weight_WSQ_w/o}
    \end{subfigure}
    \hfill
    \begin{subfigure}[b]{0.48\linewidth}
    \centering
        {\includegraphics[width=1\linewidth]{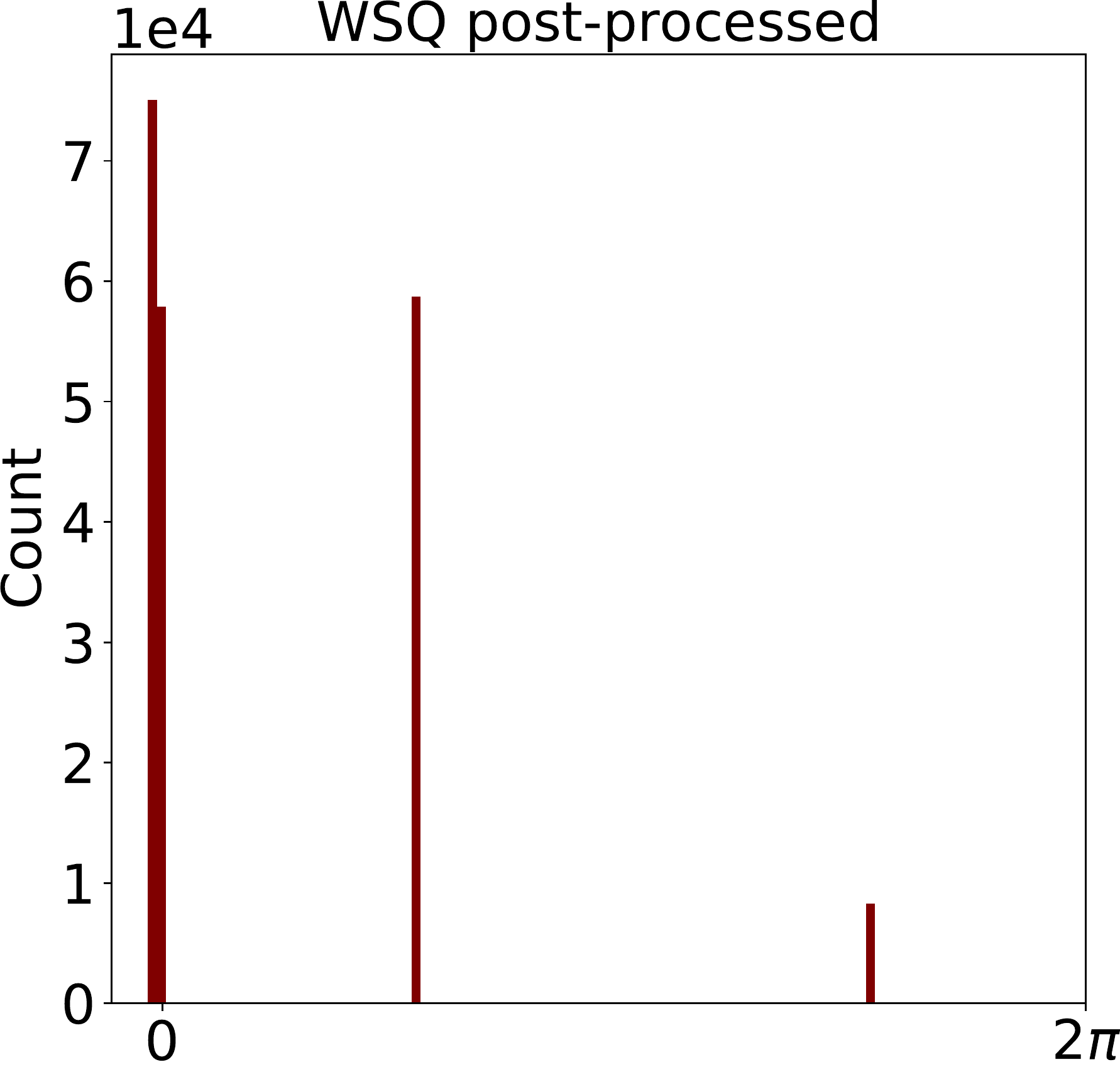}}
        \caption{WSQ w post-processing.}
        \label{fig:weight_WSQ_w}
    \end{subfigure}
    \hfill
    \begin{subfigure}[b]{0.48\linewidth}
    \centering
        {\includegraphics[width=1\linewidth]{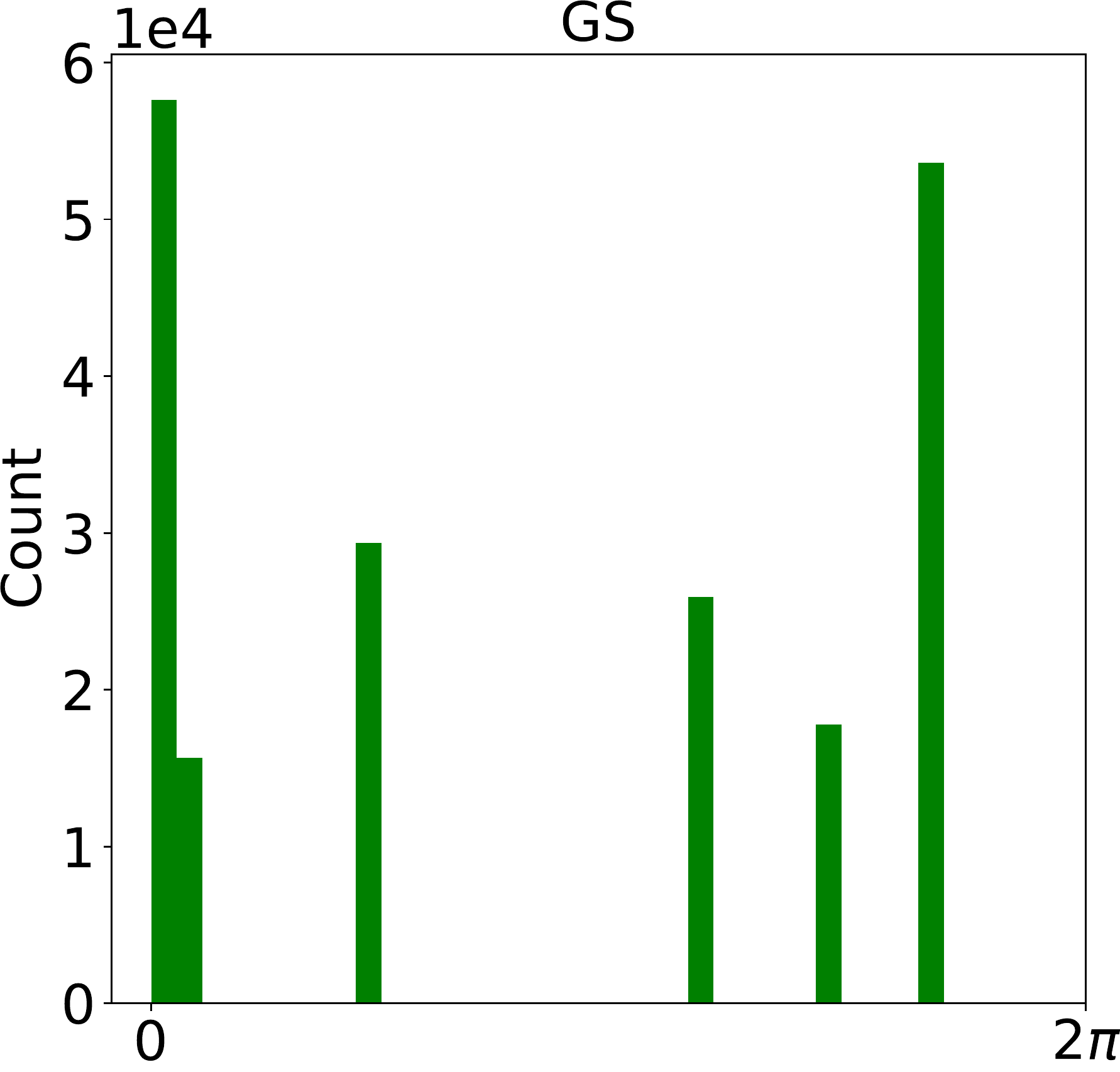}}
        \caption{Gumbel-Softmax (GS).}
        \label{fig:weight_GS}
    \end{subfigure}
    \caption{Weights distributions for the same 5-layer DONN model quantized with different quantization methods.}
    \vspace{-5mm}
    \label{fig:dist_ana}
\end{figure}
\begin{figure*}
    \centering
    \includegraphics[width=1 \linewidth]{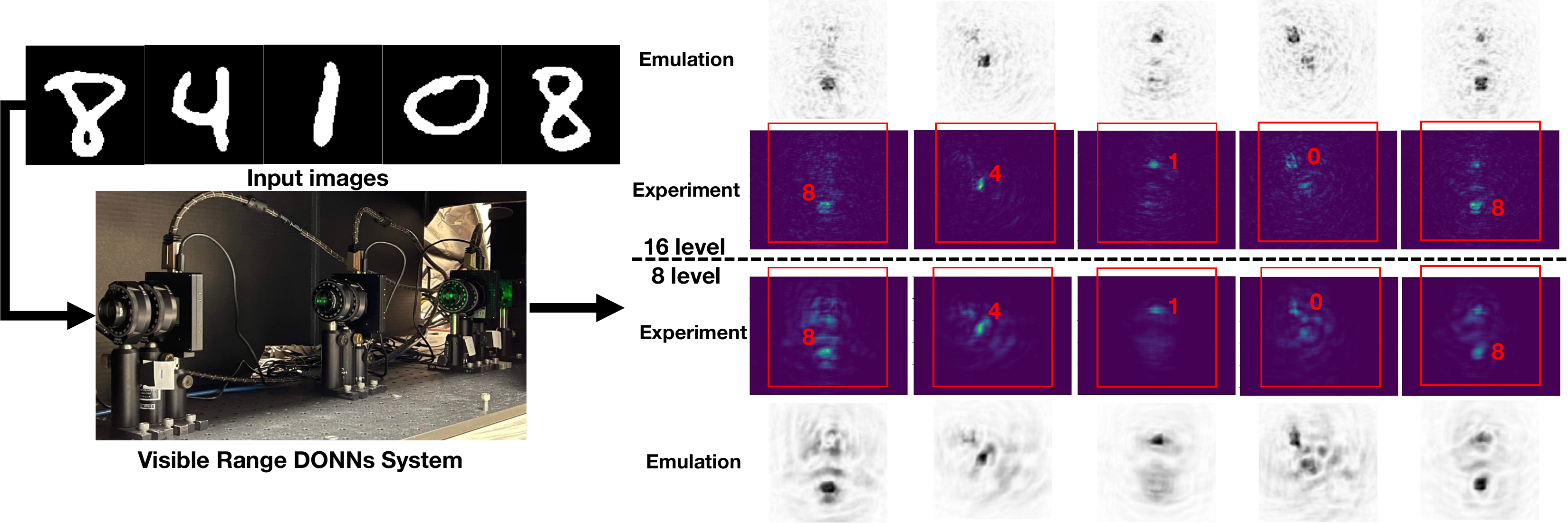}
    \caption{Experiments conducted on our existing 3-layer reconfigurable DONNs system \cite{chen2022physics} trained with our proposed approach.}
    \vspace{-3mm}
    \label{fig:prop_exp}
\end{figure*}
\vspace{-3mm}
%As we can see from table \ref{table:acc_temp}, for both MNIST and FMNIST, the system performs better with simulated annealing temperature schedule implemented in Gumbel-Softmax than the fixed temperature setups in accuracy. To be more specific, in figure \ref{fig:temp}, we show the training loss and test loss result with \texttt{log} applied for the system implemented with different temperature schedules. We can conclude from the results that: \textbf{(1)} When simulated annealing applied in the system with Gumbel-Softmax, it will preform better than that with fixed temperature implemented. With bad fixed temperature implemented (e.g., T=50 and T=1 for MNIST dataset and T=1 for FMNIST dataset), the system can get worse as training efforts increases.
%\textbf{(2)} The \textbf{even-decay} temperature schedule works best for our system as the training loss for even-decay is lowest for both datasets. In other experiments, we set even-decay temperature schedule as default for Gumbel-Softmax.
{
\vspace{-1mm}
\subsection{Comparisons with Quantization Methods}
\label{sec:result_comp}

{%In this section, we measured 8, 12, and 16 discrete data points with hardware devices that can be used to build DONN system. Moreover, we compared our work with the same system setup but implemented with fitted curve and conventional quantization method, i.e., vanilla post-training quantization (PTQ) (\cite{}), vanilla quantization-aware training (QAT) (\cite{}). {For the system implemented with conventional quantization methods, a differentiable curve will be fitted with 8, 12, 16 calibrated data points respectively. (Cunxi: i feel we might need to explain the details how we setup the PTQ and QAT. I can add this if you are not confident about this)} The model will be trained with the fitting curve, and in inference, the phase modulation controlled by voltage levels in the system will be quantized to 8, 12 and 16 respectively based on the fitting curve.

To demonstrate the advantages of the proposed GS-based training approach over existing methods, we compare the accuracy performance with existing approaches used in experimental optical studies \cite{lin2018all, zhou2021large} and various quantization methods. The targeted optical device is the same 8-level device. %Specifically, we experimentally measured discrete values of amplitude and phase modulation realized in the optical setups discussed in \textit{System Overview}. 
%The models evaluated in Figure \ref{fig:quan_vs_gumble} are trained with the proposed {GS} training framework by feeding in these measured device parameters. 
{Note that the proposed GS-based training framework is fully aware of the discrete distribution on-the-fly in the training process, in which the weights are ready for hardware deployment directly.} The trained discrete values are used to configure the device, which produces the complex-valued property of optical devices.

The baseline results are generated with various quantization methods under the same optical system setup, including post-training quantization (PTQ) \cite{krishnamoorthi2018quantizing}, quantization-aware training (QAT) \cite{krishnamoorthi2018quantizing}, and the weights sharing quantization (WSQ) \cite{han2015deep, ullrich2017soft} {shown in Figure \ref{fig:quan_vs_gumble}}. {For all the quantization methods compared, we first fit a multi-polynomial regression model that takes supply voltage as input and produces phase and amplitude value. The discrete values are selected w.r.t specific optical device responses. 
For PTQ, we train the DONNs while considering the device voltage values being feasible in \texttt{float32} precision and round the voltages to the nearest available discrete points (8, 12, or 16 discrete integers) after entire training process. {For QAT, instead of quantization after all training iterations, the values are rounded to the nearest available discrete values during the training process ({clipping over minibatch updates}), and the loss is calculated with the {clipped SLMs voltages}. Specifically, during the training of the entire dataset (one epoch), {we {explore} various options for minibatch clipping, i.e., for the MNIST training dataset with 60,000 samples, we do the quantization every 12,000 samples, 15,000 samples, 20,000 samples, 30,000 samples, and 60,000 samples, respectively. Our experimental results demonstrate that QAT training set up with one quantization per epoch over the entire dataset performs the best, which is included in Figure \ref{fig:quan_vs_gumble}.}}  
For WSQ, the input model is a pre-trained DONNs in \texttt{float32} precision, and is then quantized with weights sharing methods conducted on \texttt{KMeans} clustering \cite{pedregosa2011scikit, han2015deep, ullrich2017soft}. {Note that the quantized clusters are automatically selected by the algorithm without full awareness of the physical device.} %where the number of clusters in KMeans is the number of targeted discrete levels. 
Thus, unlike GS method, which is fully aware of the physical feasibility of the devices, WSQ method requires further post-processing to exactly match the discrete levels available on the physical devices. We provide additional discussion on the physical feasibility of devices during quantization with weight distribution analysis shown in Figure \ref{fig:dist_ana}.}  %The multi-polynomial functions are fitted fitting curve, and in inference, the phase modulation controlled by voltage levels in the system will be quantized to 8, 12 and 16 respectively based on the fitting curve.

As shown in Figure \ref{fig:quan_vs_gumble}, four different approaches are evaluated with 8, 12, and 16 discrete levels in 3-, 4-, 5-layer DONNs systems, respectively. %Note that unlike conventional digital quantization dealing with regular number representations (e.g., INT4, INT8) over full range, the discrete values in optical devices are all positive numbers (e.g., Amp and Phase curves in Figure \ref{fig:GS_curve}). 
Note that the discrete values in optical devices are all non-negative numbers. Thus, models trained and quantized with WSQ algorithm require post-processing to round all negative parameters to the nearest non-negative available value. To make comprehensive comparison, we provide both quantization results {before and after} post-processing. %To make comprehensive comparison, we provide both quantization \fixme{before and after} post-processing and without post-processing for conventional quantization algorithms. 
%Specifically, for PTQ and QAT, the quantization without post-processing will quantize the model with specific discrete levels covering both negative and positive numbers calculated from the corresponding fitting curve; the quantization with post-processing will quantize the model with available positive discrete levels calculated from the corresponding fitting curve for the device.
We can see that models trained with GS method always perform at its best accuracy (0.98 for MNIST, 0.89 for FMNIST). Specifically, \textbf{(1)} within the same depth of DONN systems, the model trained with the proposed GS-based framework demonstrates better accuracy in all cases. More importantly, for the system implemented with different discrete levels, we can see that the proposed framework is able to train the DONNs to match the best accuracy regardless of its complexity. \textbf{(2)} For systems with different structural complexity, GS method shows significant advantages in training, especially devices with fewer (e.g., 8-level) discrete levels. \textbf{(3)} For compared quantization methods, even though they work decently before post-processing, the performance will significantly degrade once post-processing is applied. For example, before post-processing, the models quantized with WSQ method perform similarly as GS method (e.g., 0.97 for MNIST, 0.88 for FMNIST). However, the accuracy after post-processing degrades more than 27\% consistently for all setups. %Similar performance degradation caused by post-processing is observed for QAT method. %For models quantized with QAT, the accuracy performance will be improved as more discrete levels are employed for the same system structure before post-processing. However, when post-processing is applied, the performance for models quantized with all three conventional quantization algorithms will degrade significantly regardless of its available discrete levels and the system complexity.
{Our results have demonstrated that the proposed differentiable discrete training framework with Gumbel-Softmax offers significant training performance improvements over conventional quantization algorithms, especially in co-designing DONN systems built with optical devices with very limited and non-uniform optical responses.} 
}

{
To understand the performance degradation by post-processing, we explore the distribution of weight parameters (Figure \ref{fig:dist_ana}) of 5-layer DONNs with 8-level devices. We can find that the performance gap between models before and after post-processing shown in Figure \ref{fig:quan_vs_gumble} comes from the unawareness of the physical optical devices, which only provides feasible positive discrete levels. 
%Specifically, the model used here is implemented with five diffractive layers and quantized with eight discrete levels. 
Figure \ref{fig:weight_full} shows the weights distribution of the model trained with full precision (\texttt{float32}), which is a zero-centered normal distribution. The distribution after WSQ quantization shown in Figure \ref{fig:weight_WSQ_w/o} follows the original distribution but results in 8 discrete levels. We can see that it is still zero-centered, covering both negative and positive discrete levels. However, the negative levels are infeasible in physical optical devices. 
Thus, after post-processing is applied, the distribution (Figure \ref{fig:weight_WSQ_w}) is significantly changed to satisfy all physical properties. All the negative discrete levels are truncated to zero and the positive discrete levels will be quantized to the nearest physically available discrete levels. Meanwhile, we can see that GS method weight distribution shown in Figure \ref{fig:weight_GS} is quite different compared to WSQ after post-processing. Note that GS method trained parameters will not need any further post-processing for hardware deployment. %is not required for hardware deployment as the algorithm is fully aware of the devices during its training process. Thus, for the model trained with the proposed GS-based framework, the weights distribute only with discrete levels in the positive range, and no significant performance degradation happens when deploying the trained model on the physical devices. 

{In conclusion}, conventional quantization methods with post-processing are applicable to discrete mapping from the device level to DONN systems but suffer from not being aware of physical feasibility for different devices. However, with the proposed GS-based framework, the model is trained with full awareness of the physical devices as discrete differential mappings. Thus, the weights acquired from the GS-based framework can be deployed directly on the physical devices without post-processing and offer significant advantages in the physics-aware algorithm-hardware correlation between numerical models and physical optical systems.

%\fixme{Missing main discussion about 1) difference between (c) and (d), and why these statistical differences are critical to the performance.}
}

}

%\subsection{Comparison with advanced compression methods \fixme{this is a new section right?}} \label{sec:comp_advance}

%{\color{blue}In this section, we try to implement our DONNs system with more advanced quantization methods in digital domain. Specifically, we transplant the quantization method in \citet{han2015deep} to fit our DONNs systems. As illustrated in \citet{han2015deep}, the pre-trained model will \textbf{(1) pruning}: be pruned and re-trained after pruning; \textbf{(2) weight-sharing}: quantized with clustering the weights and retrained with the quantized weights; \textbf{(3) encoding}: encoded weights and index with huffman encoding. However, when implemented in DONNs system, steps in this quantization method aiming at reducing the memory usage can be eliminated. As discussed in Section \ref{sec:background}, in the DONN system, the information is encoded with the input laser beam, the data movement and calculation process will happen naturally with light propagation. Thus, the intermediate results will not be stored and no memory storage is needed in a physical DONNs system. In our implementation, we only applied the first two steps, i.e., pruning and weight-sharing, to the DONNs system. We first train the DONN system without Gumbel-Softmax with learning rate 0.05 for 100 epochs as the pre-trained model, then apply pruning and weight-sharing following \citet{han2015deep}.
%The results are shown in Figure \ref{}, }

\vspace{-2mm}
\subsection{Complex-valued Confidence Evaluation}
\label{sec:result_regu}

We first evaluate the same DONN system structure trained with our GS-based method and the state-of-the-art DONN training methods in \cite{lin2018all, zhou2021large}. Specifically, we aim to overcome the training limitations stated in \cite{lin2018all} w.r.t physics limitations of shallow DONNs ($\leq 3$ layers). Second, we provide the confidence evaluation for DONN systems with different structure complexity, all shown in Table \ref{table:regu_layer}.

First, as shown in the first column and the second column in Table \ref{table:regu_layer}, DONN implemented with different diffractive layers can be trained to achieve similar high prediction accuracy (0.98 for MNIST, 0.89 for FMNIST) using our framework. For the DONN systems trained in \cite{lin2018all}, the accuracy performance decreases as the DONN system structured with less diffractive layers, in which \citet{lin2018all} stated that it was due to the fundamental limitations of optical physics in shallow DONNs. However, our proposed GS training framework has successfully boosted the prediction performance in those shallow architectures. %which is a significant boost to the state-of-the-art results discussed by \cite{lin2018all}, \cite{zhou2021large}. 
%\vspace{-3mm}
\begin{table}[!htb]
\small
\caption{Confidence evaluations of DONNs trained with 8 discrete values in the optical devices, with testing accuracy.}
\begin{tabular}{|l|l|l|l|l|l|l|}
\hline
\textbf{Dataset} & \textbf{Depths} & \textbf{\cite{lin2018all,zhou2021large}(0\%)} & \textbf{0 \%} & \textbf{1 \%} & \textbf{3 \%} & \textbf{5 \%}  \\ \hline
\multirow{3}{*}{MNIST} & \textit{D=1} & 0.670 & 0.960 & 0.398 & 0 & 0 \\ \cline{2-7} 
 & \textit{D=3} & 0.910 & 0.978 & 0.961 & 0.876 & 0.661  \\ \cline{2-7} 
 & \textit{D=5} & 0.950 & 0.979 & 0.979 & 0.979 & 0.977  \\ \hline
\multirow{3}{*}{FMNIST} & \textit{D=1} & 0.540 & 0.874 & 0.340 & 0.001 & 0  \\ \cline{2-7} 
 & \textit{D=3} & 0.830 & 0.889 & 0.791 & 0.518 & 0.278  \\ \cline{2-7} 
 & \textit{D=5} & 0.870 & 0.890 & 0.889 & 0.886 & 0.883  \\ \hline
\end{tabular}
\label{table:regu_layer}
\end{table}
%\vspace{-3mm}

While the accuracy in shallow DONNs is significantly improved, we further analyze the performance robustness of the DONNs implemented with different depth of diffractive layers. Specifically, we explore the confidence of the predictions acquired by the system. When the sample is classified correctly, we decrease the highest probability generated by softmax function (softmax of intensity values collected in the ten detector regions) by 1\%, 3\% and 5\%, and then evenly distribute to the other nine outputs, i.e., increasing the probabilities of the other outputs by 0.11\%, 0.33\% and 0.55\%, respectively. We can see that for both datasets, as the depth of DONNs increases, the prediction confidence increases, i.e., the prediction accuracy drops less w.r.t the applied errors. For example, there is almost no accuracy degradation on five-layer DONNs for MNIST, and less than 1\% degradation on FMNIST with up to 5\% applied error. However, for single-layer DONNs, the accuracy drops 56\% for MNIST and 53\% for FMNIST when 1\% error applied and drops to 0 when applied error increases to 3\% and 5\%. The confidence studies have post great intuitions in building real-world DONN systems in different device/system precision scenarios. For example, for high precision and expensive devices and camera, shallow DONNs is sufficient; while for low precision optical system, deeper DONNs are needed for robust performance.
% \textbf{As a result, it will be more difficult to deploy a less complex system on hardware as it will be vulnerable to hardware noise which is unavoidable in practical system.}
%This study provides evidences about As a result, it will be more difficult to deploy a less complex system on hardware as it will be vulnerable to hardware noise which is unavoidable in practical system.

\vspace{-2mm}
\subsection{Physical Experimental Evaluations}
\label{sec:experiment_results}

{The existing hardware system for verifying the demonstrated simulation results from GS-framework is shown in Figure \ref{fig:prop_exp} and has been presented by Chen et. al \cite{chen2022physics}. Briefly, the input images are generated at $532\,$nm wavelength laser and have the size $100\times100$. The distance between SLMs and between the last SLM and camera is set as $27.94\,$cm ($11$ inch). The final diffraction pattern is captured on a CMOS camera. For a 3-layer DONN system for MNIST data shown in Table \ref{table:regu_layer}, the trained discrete weights (voltage level) of the 3-layer DONNs shown in Table \ref{table:regu_layer} are directly deployed to the system that controls the applied phase modulation of each device, whose functionality follows the phase and amp function curves in Figure \ref{fig:GS_curve}. As shown in Figure \ref{fig:prop_exp}, we can see that the DONNs simulation results match the experimental measurements very well, which demonstrate the effectiveness of our GS framework, especially at low finite bit precision.}

\section{Conclusion}
This work studies a novel flexible device-to-system hardware-software codesign framework which enables efficient training of DONNs systems implemented with arbitrary experimental measured optical devices across the layers. Specifically, this framework realizes backpropagation through discrete parameters via Gumbel-Softmax. Our simulation results demonstrate that the DONN system optimized with the proposed framework will acquire tremendous accuracy improvements compared to the state-of-the-art quantization methods. Moreover, exploration for temperature schedule for Gumbel-Softmax in DONN system, confidence evaluation for DONN systems implemented with different layers, algorithm verification on physical optical systems are comprehensively discussed.

\noindent
\textbf{Acknowledgement:} This work is supported by University of Utah start-up fund and NSF-2019336 and NSF-2008144.
\newpage

\bibliographystyle{ACM-Reference-Format}
\bibliography{main.bib}


\begin{thebibliography}{00}

%%% the Web address.

\ifx \showCODEN    \undefined \def \showCODEN     #1{\unskip}     \fi
\ifx \showDOI      \undefined \def \showDOI       #1{#1}\fi
\ifx \showISBNx    \undefined \def \showISBNx     #1{\unskip}     \fi
\ifx \showISBNxiii \undefined \def \showISBNxiii  #1{\unskip}     \fi
\ifx \showISSN     \undefined \def \showISSN      #1{\unskip}     \fi
\ifx \showLCCN     \undefined \def \showLCCN      #1{\unskip}     \fi
\ifx \shownote     \undefined \def \shownote      #1{#1}          \fi
\ifx \showarticletitle \undefined \def \showarticletitle #1{#1}   \fi
\ifx \showURL      \undefined \def \showURL       {\relax}        \fi
% invisible to TeX
\providecommand\bibfield[2]{#2}
\providecommand\bibinfo[2]{#2}
\providecommand\natexlab[1]{#1}
\providecommand\showeprint[2][]{arXiv:#2}

\bibitem[\protect\citeauthoryear{Abadi, Barham, Chen, Chen, Davis, Dean, Devin,
  Ghemawat, Irving, Isard, et~al\mbox{.}}{Abadi et~al\mbox{.}}{2016}]%
        {abadi2016tensorflow}
\bibfield{author}{\bibinfo{person}{Mart{\'\i}n Abadi}, \bibinfo{person}{Paul
  Barham}, \bibinfo{person}{Jianmin Chen}, \bibinfo{person}{Zhifeng Chen},
  \bibinfo{person}{Andy Davis}, \bibinfo{person}{Jeffrey Dean},
  \bibinfo{person}{Matthieu Devin}, \bibinfo{person}{Sanjay Ghemawat},
  \bibinfo{person}{Geoffrey Irving}, \bibinfo{person}{Michael Isard},
  {et~al\mbox{.}}} \bibinfo{year}{2016}\natexlab{}.
\newblock \showarticletitle{{Tensorflow: A system for large-scale Machine
  Learning}}.
\newblock \bibinfo{journal}{{\em Symp. on Operating System Design and
  Implementation (OSDI)\/}} (\bibinfo{year}{2016}), \bibinfo{pages}{265--283}.
\newblock


\bibitem[\protect\citeauthoryear{Baevski, Schneider, and Auli}{Baevski
  et~al\mbox{.}}{2019}]%
        {baevski2019vq}
\bibfield{author}{\bibinfo{person}{Alexei Baevski}, \bibinfo{person}{Steffen
  Schneider}, {and} \bibinfo{person}{Michael Auli}.}
  \bibinfo{year}{2019}\natexlab{}.
\newblock \showarticletitle{vq-wav2vec: Self-supervised learning of discrete
  speech representations}.
\newblock \bibinfo{journal}{{\em arXiv preprint arXiv:1910.05453\/}}
  (\bibinfo{year}{2019}).
\newblock


\bibitem[\protect\citeauthoryear{Chen, Li, Lou, Fan, Tang, Sensale-Rodriguez,
  Yu, and Gao}{Chen et~al\mbox{.}}{2022a}]%
        {chen2022physics}
\bibfield{author}{\bibinfo{person}{Ruiyang Chen}, \bibinfo{person}{Yingjie Li},
  \bibinfo{person}{Minhan Lou}, \bibinfo{person}{Jichao Fan},
  \bibinfo{person}{Yingheng Tang}, \bibinfo{person}{Berardi Sensale-Rodriguez},
  \bibinfo{person}{Cunxi Yu}, {and} \bibinfo{person}{Weilu Gao}.}
  \bibinfo{year}{2022}\natexlab{a}.
\newblock \showarticletitle{Physics-aware Complex-valued Adversarial Machine
  Learning in Reconfigurable Diffractive All-optical Neural Network}.
\newblock \bibinfo{journal}{{\em arXiv preprint arXiv:2203.06055\/}}
  (\bibinfo{year}{2022}).
\newblock


\bibitem[\protect\citeauthoryear{Chen, Li, Lou, Yu, and Gao}{Chen
  et~al\mbox{.}}{2022b}]%
        {chen2022complex}
\bibfield{author}{\bibinfo{person}{Ruiyang Chen}, \bibinfo{person}{Yingjie Li},
  \bibinfo{person}{Minhan Lou}, \bibinfo{person}{Cunxi Yu}, {and}
  \bibinfo{person}{Weilu Gao}.} \bibinfo{year}{2022}\natexlab{b}.
\newblock \showarticletitle{Complex-valued Reconfigurable Diffractive Optical
  Neural Networks using Cost-effective Spatial Light Modulators}. In
  \bibinfo{booktitle}{{\em CLEO: Applications and Technology}}. Optica
  Publishing Group, \bibinfo{pages}{JTh3B--56}.
\newblock


\bibitem[\protect\citeauthoryear{Ersoy}{Ersoy}{2006}]%
        {ersoy2006diffraction}
\bibfield{author}{\bibinfo{person}{Okan~K Ersoy}.}
  \bibinfo{year}{2006}\natexlab{}.
\newblock \bibinfo{booktitle}{{\em Diffraction, Fourier optics and imaging}}.
\newblock \bibinfo{publisher}{John Wiley \& Sons}.
\newblock


\bibitem[\protect\citeauthoryear{Feldmann, Youngblood, Wright, Bhaskaran, and
  Pernice}{Feldmann et~al\mbox{.}}{2019}]%
        {feldmann2019all}
\bibfield{author}{\bibinfo{person}{J Feldmann}, \bibinfo{person}{N Youngblood},
  \bibinfo{person}{C~David Wright}, \bibinfo{person}{H Bhaskaran}, {and}
  \bibinfo{person}{WHP Pernice}.} \bibinfo{year}{2019}\natexlab{}.
\newblock \showarticletitle{All-optical spiking neurosynaptic networks with
  self-learning capabilities}.
\newblock \bibinfo{journal}{{\em Nature\/}} \bibinfo{volume}{569},
  \bibinfo{number}{7755} (\bibinfo{year}{2019}), \bibinfo{pages}{208--214}.
\newblock


\bibitem[\protect\citeauthoryear{Fu, Zhang, Li, Yu, and Lin}{Fu
  et~al\mbox{.}}{2021a}]%
        {fu2021a3c}
\bibfield{author}{\bibinfo{person}{Yonggan Fu}, \bibinfo{person}{Yongan Zhang},
  \bibinfo{person}{Chaojian Li}, \bibinfo{person}{Zhongzhi Yu}, {and}
  \bibinfo{person}{Yingyan Lin}.} \bibinfo{year}{2021}\natexlab{a}.
\newblock \showarticletitle{A3C-S: Automated Agent Accelerator Co-Search
  towards Efficient Deep Reinforcement Learning}. In \bibinfo{booktitle}{{\em
  2021 58th ACM/IEEE Design Automation Conference (DAC)}}. IEEE,
  \bibinfo{pages}{13--18}.
\newblock


\bibitem[\protect\citeauthoryear{Fu, Zhang, Zhang, Cox, and Lin}{Fu
  et~al\mbox{.}}{2021b}]%
        {fu2021auto}
\bibfield{author}{\bibinfo{person}{Yonggan Fu}, \bibinfo{person}{Yongan Zhang},
  \bibinfo{person}{Yang Zhang}, \bibinfo{person}{David Cox}, {and}
  \bibinfo{person}{Yingyan Lin}.} \bibinfo{year}{2021}\natexlab{b}.
\newblock \showarticletitle{Auto-NBA: Efficient and effective search over the
  joint space of networks, bitwidths, and accelerators}. In
  \bibinfo{booktitle}{{\em International Conference on Machine Learning}}.
  PMLR, \bibinfo{pages}{3505--3517}.
\newblock


\bibitem[\protect\citeauthoryear{Gao, Yu, and Chen}{Gao et~al\mbox{.}}{2021a}]%
        {gao2021artificial}
\bibfield{author}{\bibinfo{person}{Weilu Gao}, \bibinfo{person}{Cunxi Yu},
  {and} \bibinfo{person}{Ruiyang Chen}.} \bibinfo{year}{2021}\natexlab{a}.
\newblock \showarticletitle{Artificial Intelligence Accelerators Based on
  Graphene Optoelectronic Devices}.
\newblock \bibinfo{journal}{{\em Advanced Photonics Research\/}}
  \bibinfo{volume}{2}, \bibinfo{number}{6} (\bibinfo{year}{2021}),
  \bibinfo{pages}{2100048}.
\newblock


\bibitem[\protect\citeauthoryear{Gao, Yu, and Chen}{Gao et~al\mbox{.}}{2021b}]%
        {gao2021graphene}
\bibfield{author}{\bibinfo{person}{Weilu Gao}, \bibinfo{person}{Cunxi Yu},
  {and} \bibinfo{person}{Ruiyang Chen}.} \bibinfo{year}{2021}\natexlab{b}.
\newblock \showarticletitle{Graphene Optoelectronic Artificial Intelligence
  Accelerators}. In \bibinfo{booktitle}{{\em CLEO: QELS\_Fundamental Science}}.
  Optical Society of America, \bibinfo{pages}{JTu3A--88}.
\newblock


\bibitem[\protect\citeauthoryear{Gu, Feng, Zhao, Ying, Chen, and Pan}{Gu
  et~al\mbox{.}}{2021a}]%
        {gu2021efficient}
\bibfield{author}{\bibinfo{person}{Jiaqi Gu}, \bibinfo{person}{Chenghao Feng},
  \bibinfo{person}{Zheng Zhao}, \bibinfo{person}{Zhoufeng Ying},
  \bibinfo{person}{Ray~T Chen}, {and} \bibinfo{person}{David~Z Pan}.}
  \bibinfo{year}{2021}\natexlab{a}.
\newblock \showarticletitle{Efficient on-chip learning for optical neural
  networks through power-aware sparse zeroth-order optimization}. In
  \bibinfo{booktitle}{{\em Proc. AAAI}}.
\newblock


\bibitem[\protect\citeauthoryear{Gu, Feng, Zhao, Ying, Liu, Chen, and Pan}{Gu
  et~al\mbox{.}}{2021b}]%
        {gu2021squeezelight}
\bibfield{author}{\bibinfo{person}{Jiaqi Gu}, \bibinfo{person}{Chenghao Feng},
  \bibinfo{person}{Zheng Zhao}, \bibinfo{person}{Zhoufeng Ying},
  \bibinfo{person}{Mingjie Liu}, \bibinfo{person}{Ray~T Chen}, {and}
  \bibinfo{person}{David~Z Pan}.} \bibinfo{year}{2021}\natexlab{b}.
\newblock \showarticletitle{SqueezeLight: towards scalable optical neural
  networks with multi-operand ring resonators}. In \bibinfo{booktitle}{{\em
  2021 Design, Automation \& Test in Europe Conference \& Exhibition (DATE)}}.
  IEEE, \bibinfo{pages}{238--243}.
\newblock


\bibitem[\protect\citeauthoryear{Gu, Zhao, Feng, Li, Chen, and Pan}{Gu
  et~al\mbox{.}}{2020a}]%
        {gu2020flops}
\bibfield{author}{\bibinfo{person}{Jiaqi Gu}, \bibinfo{person}{Zheng Zhao},
  \bibinfo{person}{Chenghao Feng}, \bibinfo{person}{Wuxi Li},
  \bibinfo{person}{Ray~T Chen}, {and} \bibinfo{person}{David~Z Pan}.}
  \bibinfo{year}{2020}\natexlab{a}.
\newblock \showarticletitle{FLOPS: efficient on-chip learning for optical
  neural networks through stochastic zeroth-order optimization}. In
  \bibinfo{booktitle}{{\em 2020 57th ACM/IEEE Design Automation Conference
  (DAC)}}. IEEE, \bibinfo{pages}{1--6}.
\newblock


\bibitem[\protect\citeauthoryear{Gu, Zhao, Feng, Liu, Chen, and Pan}{Gu
  et~al\mbox{.}}{2020b}]%
        {gu2020towards}
\bibfield{author}{\bibinfo{person}{Jiaqi Gu}, \bibinfo{person}{Zheng Zhao},
  \bibinfo{person}{Chenghao Feng}, \bibinfo{person}{Mingjie Liu},
  \bibinfo{person}{Ray~T Chen}, {and} \bibinfo{person}{David~Z Pan}.}
  \bibinfo{year}{2020}\natexlab{b}.
\newblock \showarticletitle{Towards area-efficient optical neural networks: an
  FFT-based architecture}. In \bibinfo{booktitle}{{\em 2020 25th Asia and South
  Pacific Design Automation Conference (ASP-DAC)}}. IEEE,
  \bibinfo{pages}{476--481}.
\newblock


\bibitem[\protect\citeauthoryear{Gumbel}{Gumbel}{1954}]%
        {gumbel1954statistical}
\bibfield{author}{\bibinfo{person}{Emil~Julius Gumbel}.}
  \bibinfo{year}{1954}\natexlab{}.
\newblock \bibinfo{booktitle}{{\em Statistical theory of extreme values and
  some practical applications: a series of lectures}}.
  Vol.~\bibinfo{volume}{33}.
\newblock \bibinfo{publisher}{US Government Printing Office}.
\newblock


\bibitem[\protect\citeauthoryear{Han, Mao, and Dally}{Han
  et~al\mbox{.}}{2015}]%
        {han2015deep}
\bibfield{author}{\bibinfo{person}{Song Han}, \bibinfo{person}{Huizi Mao},
  {and} \bibinfo{person}{William~J Dally}.} \bibinfo{year}{2015}\natexlab{}.
\newblock \showarticletitle{{Deep compression: Compressing Deep Neural Networks
  with Pruning, Trained Quantization and Huffman Coding}}.
\newblock \bibinfo{journal}{{\em Advances in Neural Information Processing
  Systems (NIPS)\/}} (\bibinfo{year}{2015}).
\newblock


\bibitem[\protect\citeauthoryear{Hazan, Ratzker, Zhang, Katiyi, Frage, Sokol,
  Gogotsi, and Karabchevsky}{Hazan et~al\mbox{.}}{2021}]%
        {hazan2021ti}
\bibfield{author}{\bibinfo{person}{Adir Hazan}, \bibinfo{person}{Barak
  Ratzker}, \bibinfo{person}{Danzhen Zhang}, \bibinfo{person}{Aviad Katiyi},
  \bibinfo{person}{Nachum Frage}, \bibinfo{person}{Maxim Sokol},
  \bibinfo{person}{Yury Gogotsi}, {and} \bibinfo{person}{Alina Karabchevsky}.}
  \bibinfo{year}{2021}\natexlab{}.
\newblock \showarticletitle{Ti $ \_3 $ C $ \_2 $ T $ \_x $ MXene Enabled
  All-Optical Nonlinear Activation Function for On-Chip Photonic Deep Neural
  Networks}.
\newblock \bibinfo{journal}{{\em arXiv preprint arXiv:2109.09177\/}}
  (\bibinfo{year}{2021}).
\newblock


\bibitem[\protect\citeauthoryear{He, Ye, Shen, and Zhang}{He
  et~al\mbox{.}}{2020}]%
        {he2020milenas}
\bibfield{author}{\bibinfo{person}{Chaoyang He}, \bibinfo{person}{Haishan Ye},
  \bibinfo{person}{Li Shen}, {and} \bibinfo{person}{Tong Zhang}.}
  \bibinfo{year}{2020}\natexlab{}.
\newblock \showarticletitle{Milenas: Efficient neural architecture search via
  mixed-level reformulation}. In \bibinfo{booktitle}{{\em Proceedings of the
  IEEE/CVF Conference on Computer Vision and Pattern Recognition}}.
  \bibinfo{pages}{11993--12002}.
\newblock


\bibitem[\protect\citeauthoryear{Jang, Gu, and Poole}{Jang
  et~al\mbox{.}}{2016}]%
        {jang2016categorical}
\bibfield{author}{\bibinfo{person}{Eric Jang}, \bibinfo{person}{Shixiang Gu},
  {and} \bibinfo{person}{Ben Poole}.} \bibinfo{year}{2016}\natexlab{}.
\newblock \showarticletitle{Categorical reparameterization with
  gumbel-softmax}.
\newblock \bibinfo{journal}{{\em arXiv preprint arXiv:1611.01144\/}}
  (\bibinfo{year}{2016}).
\newblock


\bibitem[\protect\citeauthoryear{Jouppi, Young, Patil, Patterson, Agrawal,
  Bajwa, Bates, Bhatia, Boden, Borchers, et~al\mbox{.}}{Jouppi
  et~al\mbox{.}}{2017}]%
        {jouppi2017datacenter}
\bibfield{author}{\bibinfo{person}{Norman~P Jouppi}, \bibinfo{person}{Cliff
  Young}, \bibinfo{person}{Nishant Patil}, \bibinfo{person}{David Patterson},
  \bibinfo{person}{Gaurav Agrawal}, \bibinfo{person}{Raminder Bajwa},
  \bibinfo{person}{Sarah Bates}, \bibinfo{person}{Suresh Bhatia},
  \bibinfo{person}{Nan Boden}, \bibinfo{person}{Al Borchers}, {et~al\mbox{.}}}
  \bibinfo{year}{2017}\natexlab{}.
\newblock \showarticletitle{{In-datacenter Performance Analysis of a Tensor
  Processing Unit}}.
\newblock \bibinfo{journal}{{\em Int'l Symp. on Computer Architecture
  (ISCA)\/}} (\bibinfo{year}{2017}), \bibinfo{pages}{1--12}.
\newblock


\bibitem[\protect\citeauthoryear{Kingma and Ba}{Kingma and Ba}{2014}]%
        {kingma2014adam}
\bibfield{author}{\bibinfo{person}{Diederik~P Kingma} {and}
  \bibinfo{person}{Jimmy Ba}.} \bibinfo{year}{2014}\natexlab{}.
\newblock \showarticletitle{{Adam: A method for Stochastic Optimization}}.
\newblock \bibinfo{journal}{{\em arXiv preprint arXiv:1412.6980\/}}
  (\bibinfo{year}{2014}).
\newblock


\bibitem[\protect\citeauthoryear{Krishnamoorthi}{Krishnamoorthi}{2018}]%
        {krishnamoorthi2018quantizing}
\bibfield{author}{\bibinfo{person}{Raghuraman Krishnamoorthi}.}
  \bibinfo{year}{2018}\natexlab{}.
\newblock \showarticletitle{Quantizing deep convolutional networks for
  efficient inference: A whitepaper}.
\newblock \bibinfo{journal}{{\em arXiv preprint arXiv:1806.08342\/}}
  (\bibinfo{year}{2018}).
\newblock


\bibitem[\protect\citeauthoryear{LeCun}{LeCun}{1998}]%
        {lecun1998mnist}
\bibfield{author}{\bibinfo{person}{Yann LeCun}.}
  \bibinfo{year}{1998}\natexlab{}.
\newblock \showarticletitle{The MNIST database of handwritten digits}.
\newblock \bibinfo{journal}{{\em http://yann. lecun. com/exdb/mnist/\/}}
  (\bibinfo{year}{1998}).
\newblock


\bibitem[\protect\citeauthoryear{LeCun, Bengio, and Hinton}{LeCun
  et~al\mbox{.}}{2015}]%
        {lecun2015deep}
\bibfield{author}{\bibinfo{person}{Yann LeCun}, \bibinfo{person}{Yoshua
  Bengio}, {and} \bibinfo{person}{Geoffrey Hinton}.}
  \bibinfo{year}{2015}\natexlab{}.
\newblock \showarticletitle{Deep learning}.
\newblock \bibinfo{journal}{{\em nature\/}} \bibinfo{volume}{521},
  \bibinfo{number}{7553} (\bibinfo{year}{2015}), \bibinfo{pages}{436--444}.
\newblock


\bibitem[\protect\citeauthoryear{Li, Yu, Zhang, Fu, and Lin}{Li
  et~al\mbox{.}}{2021b}]%
        {li2021has}
\bibfield{author}{\bibinfo{person}{Mengquan Li}, \bibinfo{person}{Zhongzhi Yu},
  \bibinfo{person}{Yongan Zhang}, \bibinfo{person}{Yonggan Fu}, {and}
  \bibinfo{person}{Yingyan Lin}.} \bibinfo{year}{2021}\natexlab{b}.
\newblock \showarticletitle{O-HAS: Optical Hardware Accelerator Search for
  Boosting Both Acceleration Performance and Development Speed}. In
  \bibinfo{booktitle}{{\em 2021 IEEE/ACM International Conference On Computer
  Aided Design (ICCAD)}}. IEEE, \bibinfo{pages}{1--9}.
\newblock


\bibitem[\protect\citeauthoryear{Li, Chen, Rodriguez, Gao, and Yu}{Li
  et~al\mbox{.}}{2021a}]%
        {li2020multi}
\bibfield{author}{\bibinfo{person}{Yingjie Li}, \bibinfo{person}{Ruiyang Chen},
  \bibinfo{person}{Berardi~Sensale Rodriguez}, \bibinfo{person}{Weilu Gao},
  {and} \bibinfo{person}{Cunxi Yu}.} \bibinfo{year}{2021}\natexlab{a}.
\newblock \showarticletitle{Multi-Task Learning in Diffractive Deep Neural
  Networks via Hardware-Software Co-design}.
\newblock \bibinfo{journal}{{\em Scientific Reports\/}} (\bibinfo{year}{2021}),
  \bibinfo{pages}{1--9}.
\newblock


\bibitem[\protect\citeauthoryear{Li and Yu}{Li and Yu}{2021}]%
        {li2021late}
\bibfield{author}{\bibinfo{person}{Yingjie Li} {and} \bibinfo{person}{Cunxi
  Yu}.} \bibinfo{year}{2021}\natexlab{}.
\newblock \showarticletitle{Late Breaking Results: Physical Adversarial Attacks
  of Diffractive Deep Neural Networks}. In \bibinfo{booktitle}{{\em 2021 58th
  ACM/IEEE Design Automation Conference (DAC)}}. IEEE,
  \bibinfo{pages}{1374--1375}.
\newblock


\bibitem[\protect\citeauthoryear{Lin, Rivenson, Yardimci, Veli, Luo, Jarrahi,
  and Ozcan}{Lin et~al\mbox{.}}{2018}]%
        {lin2018all}
\bibfield{author}{\bibinfo{person}{Xing Lin}, \bibinfo{person}{Yair Rivenson},
  \bibinfo{person}{Nezih~T Yardimci}, \bibinfo{person}{Muhammed Veli},
  \bibinfo{person}{Yi Luo}, \bibinfo{person}{Mona Jarrahi}, {and}
  \bibinfo{person}{Aydogan Ozcan}.} \bibinfo{year}{2018}\natexlab{}.
\newblock \showarticletitle{All-optical machine learning using diffractive deep
  neural networks}.
\newblock \bibinfo{journal}{{\em Science\/}} \bibinfo{volume}{361},
  \bibinfo{number}{6406} (\bibinfo{year}{2018}), \bibinfo{pages}{1004--1008}.
\newblock


\bibitem[\protect\citeauthoryear{Maddison, Mnih, and Teh}{Maddison
  et~al\mbox{.}}{2016}]%
        {maddison2016concrete}
\bibfield{author}{\bibinfo{person}{Chris~J Maddison}, \bibinfo{person}{Andriy
  Mnih}, {and} \bibinfo{person}{Yee~Whye Teh}.}
  \bibinfo{year}{2016}\natexlab{}.
\newblock \showarticletitle{The concrete distribution: A continuous relaxation
  of discrete random variables}.
\newblock \bibinfo{journal}{{\em arXiv preprint arXiv:1611.00712\/}}
  (\bibinfo{year}{2016}).
\newblock


\bibitem[\protect\citeauthoryear{Mengu, Luo, Rivenson, and Ozcan}{Mengu
  et~al\mbox{.}}{2019}]%
        {mengu2019analysis}
\bibfield{author}{\bibinfo{person}{Deniz Mengu}, \bibinfo{person}{Yi Luo},
  \bibinfo{person}{Yair Rivenson}, {and} \bibinfo{person}{Aydogan Ozcan}.}
  \bibinfo{year}{2019}\natexlab{}.
\newblock \showarticletitle{Analysis of diffractive optical neural networks and
  their integration with electronic neural networks}.
\newblock \bibinfo{journal}{{\em IEEE Journal of Selected Topics in Quantum
  Electronics\/}} \bibinfo{volume}{26}, \bibinfo{number}{1}
  (\bibinfo{year}{2019}), \bibinfo{pages}{1--14}.
\newblock


\bibitem[\protect\citeauthoryear{Mengu, Rivenson, and Ozcan}{Mengu
  et~al\mbox{.}}{2020}]%
        {mengu2020scale}
\bibfield{author}{\bibinfo{person}{Deniz Mengu}, \bibinfo{person}{Yair
  Rivenson}, {and} \bibinfo{person}{Aydogan Ozcan}.}
  \bibinfo{year}{2020}\natexlab{}.
\newblock \showarticletitle{Scale-, shift-and rotation-invariant diffractive
  optical networks}.
\newblock \bibinfo{journal}{{\em arXiv preprint arXiv:2010.12747\/}}
  (\bibinfo{year}{2020}).
\newblock


\bibitem[\protect\citeauthoryear{Pedregosa, Varoquaux, Gramfort, Michel,
  Thirion, Grisel, Blondel, Prettenhofer, Weiss, Dubourg,
  et~al\mbox{.}}{Pedregosa et~al\mbox{.}}{2011}]%
        {pedregosa2011scikit}
\bibfield{author}{\bibinfo{person}{Fabian Pedregosa}, \bibinfo{person}{Ga{\"e}l
  Varoquaux}, \bibinfo{person}{Alexandre Gramfort}, \bibinfo{person}{Vincent
  Michel}, \bibinfo{person}{Bertrand Thirion}, \bibinfo{person}{Olivier
  Grisel}, \bibinfo{person}{Mathieu Blondel}, \bibinfo{person}{Peter
  Prettenhofer}, \bibinfo{person}{Ron Weiss}, \bibinfo{person}{Vincent
  Dubourg}, {et~al\mbox{.}}} \bibinfo{year}{2011}\natexlab{}.
\newblock \showarticletitle{Scikit-learn: Machine learning in Python}.
\newblock \bibinfo{journal}{{\em the Journal of machine Learning research\/}}
  \bibinfo{volume}{12} (\bibinfo{year}{2011}), \bibinfo{pages}{2825--2830}.
\newblock


\bibitem[\protect\citeauthoryear{Rahman, Li, Mengu, Rivenson, and Ozcan}{Rahman
  et~al\mbox{.}}{2020}]%
        {rahman2020ensemble}
\bibfield{author}{\bibinfo{person}{Md~Sadman~Sakib Rahman},
  \bibinfo{person}{Jingxi Li}, \bibinfo{person}{Deniz Mengu},
  \bibinfo{person}{Yair Rivenson}, {and} \bibinfo{person}{Aydogan Ozcan}.}
  \bibinfo{year}{2020}\natexlab{}.
\newblock \showarticletitle{Ensemble learning of diffractive optical networks}.
\newblock \bibinfo{journal}{{\em arXiv preprint arXiv:2009.06869\/}}
  (\bibinfo{year}{2020}).
\newblock


\bibitem[\protect\citeauthoryear{Senior, Evans, Jumper, Kirkpatrick, Sifre,
  Green, Qin, {\v{Z}}{\'\i}dek, Nelson, Bridgland, et~al\mbox{.}}{Senior
  et~al\mbox{.}}{2020}]%
        {senior2020improved}
\bibfield{author}{\bibinfo{person}{Andrew~W Senior}, \bibinfo{person}{Richard
  Evans}, \bibinfo{person}{John Jumper}, \bibinfo{person}{James Kirkpatrick},
  \bibinfo{person}{Laurent Sifre}, \bibinfo{person}{Tim Green},
  \bibinfo{person}{Chongli Qin}, \bibinfo{person}{Augustin {\v{Z}}{\'\i}dek},
  \bibinfo{person}{Alexander~WR Nelson}, \bibinfo{person}{Alex Bridgland},
  {et~al\mbox{.}}} \bibinfo{year}{2020}\natexlab{}.
\newblock \showarticletitle{Improved protein structure prediction using
  potentials from deep learning}.
\newblock \bibinfo{journal}{{\em Nature\/}} \bibinfo{volume}{577},
  \bibinfo{number}{7792} (\bibinfo{year}{2020}), \bibinfo{pages}{706--710}.
\newblock


\bibitem[\protect\citeauthoryear{Sharma, Park, Amaro, Thwaites, Kotha, Gupta,
  Kim, Mishra, and Esmaeilzadeh}{Sharma et~al\mbox{.}}{2016}]%
        {sharma2016dnnweaver}
\bibfield{author}{\bibinfo{person}{Hardik Sharma}, \bibinfo{person}{Jongse
  Park}, \bibinfo{person}{Emmanuel Amaro}, \bibinfo{person}{Bradley Thwaites},
  \bibinfo{person}{Praneetha Kotha}, \bibinfo{person}{Anmol Gupta},
  \bibinfo{person}{Joon~Kyung Kim}, \bibinfo{person}{Asit Mishra}, {and}
  \bibinfo{person}{Hadi Esmaeilzadeh}.} \bibinfo{year}{2016}\natexlab{}.
\newblock \showarticletitle{Dnnweaver: From high-level deep network models to
  fpga acceleration}. In \bibinfo{booktitle}{{\em the Workshop on Cognitive
  Architectures}}.
\newblock


\bibitem[\protect\citeauthoryear{Shen, Harris, Skirlo, Prabhu, Baehr-Jones,
  Hochberg, Sun, Zhao, Larochelle, Englund, et~al\mbox{.}}{Shen
  et~al\mbox{.}}{2017}]%
        {shen2017deep}
\bibfield{author}{\bibinfo{person}{Yichen Shen}, \bibinfo{person}{Nicholas~C
  Harris}, \bibinfo{person}{Scott Skirlo}, \bibinfo{person}{Mihika Prabhu},
  \bibinfo{person}{Tom Baehr-Jones}, \bibinfo{person}{Michael Hochberg},
  \bibinfo{person}{Xin Sun}, \bibinfo{person}{Shijie Zhao},
  \bibinfo{person}{Hugo Larochelle}, \bibinfo{person}{Dirk Englund},
  {et~al\mbox{.}}} \bibinfo{year}{2017}\natexlab{}.
\newblock \showarticletitle{Deep learning with coherent nanophotonic circuits}.
\newblock \bibinfo{journal}{{\em Nature Photonics\/}} \bibinfo{volume}{11},
  \bibinfo{number}{7} (\bibinfo{year}{2017}), \bibinfo{pages}{441}.
\newblock


\bibitem[\protect\citeauthoryear{Silver, Schrittwieser, Simonyan, Antonoglou,
  Huang, Guez, Hubert, Baker, Lai, Bolton, et~al\mbox{.}}{Silver
  et~al\mbox{.}}{2017}]%
        {silver2017mastering}
\bibfield{author}{\bibinfo{person}{David Silver}, \bibinfo{person}{Julian
  Schrittwieser}, \bibinfo{person}{Karen Simonyan}, \bibinfo{person}{Ioannis
  Antonoglou}, \bibinfo{person}{Aja Huang}, \bibinfo{person}{Arthur Guez},
  \bibinfo{person}{Thomas Hubert}, \bibinfo{person}{Lucas Baker},
  \bibinfo{person}{Matthew Lai}, \bibinfo{person}{Adrian Bolton},
  {et~al\mbox{.}}} \bibinfo{year}{2017}\natexlab{}.
\newblock \showarticletitle{Mastering the game of go without human knowledge}.
\newblock \bibinfo{journal}{{\em nature\/}} \bibinfo{volume}{550},
  \bibinfo{number}{7676} (\bibinfo{year}{2017}), \bibinfo{pages}{354--359}.
\newblock


\bibitem[\protect\citeauthoryear{Tait, De~Lima, Zhou, Wu, Nahmias, Shastri, and
  Prucnal}{Tait et~al\mbox{.}}{2017}]%
        {tait2017neuromorphic}
\bibfield{author}{\bibinfo{person}{Alexander~N Tait},
  \bibinfo{person}{Thomas~Ferreira De~Lima}, \bibinfo{person}{Ellen Zhou},
  \bibinfo{person}{Allie~X Wu}, \bibinfo{person}{Mitchell~A Nahmias},
  \bibinfo{person}{Bhavin~J Shastri}, {and} \bibinfo{person}{Paul~R Prucnal}.}
  \bibinfo{year}{2017}\natexlab{}.
\newblock \showarticletitle{Neuromorphic photonic networks using silicon
  photonic weight banks}.
\newblock \bibinfo{journal}{{\em Scientific reports\/}} \bibinfo{volume}{7},
  \bibinfo{number}{1} (\bibinfo{year}{2017}), \bibinfo{pages}{1--10}.
\newblock


\bibitem[\protect\citeauthoryear{Ullrich, Meeds, and Welling}{Ullrich
  et~al\mbox{.}}{2017}]%
        {ullrich2017soft}
\bibfield{author}{\bibinfo{person}{Karen Ullrich}, \bibinfo{person}{Edward
  Meeds}, {and} \bibinfo{person}{Max Welling}.}
  \bibinfo{year}{2017}\natexlab{}.
\newblock \showarticletitle{Soft weight-sharing for neural network
  compression}.
\newblock \bibinfo{journal}{{\em arXiv preprint arXiv:1702.04008\/}}
  (\bibinfo{year}{2017}).
\newblock


\bibitem[\protect\citeauthoryear{Wu, Dai, Zhang, Wang, Sun, Wu, Tian, Vajda,
  Jia, and Keutzer}{Wu et~al\mbox{.}}{2019}]%
        {wu2019fbnet}
\bibfield{author}{\bibinfo{person}{Bichen Wu}, \bibinfo{person}{Xiaoliang Dai},
  \bibinfo{person}{Peizhao Zhang}, \bibinfo{person}{Yanghan Wang},
  \bibinfo{person}{Fei Sun}, \bibinfo{person}{Yiming Wu},
  \bibinfo{person}{Yuandong Tian}, \bibinfo{person}{Peter Vajda},
  \bibinfo{person}{Yangqing Jia}, {and} \bibinfo{person}{Kurt Keutzer}.}
  \bibinfo{year}{2019}\natexlab{}.
\newblock \showarticletitle{Fbnet: Hardware-aware efficient convnet design via
  differentiable neural architecture search}. In \bibinfo{booktitle}{{\em
  Proceedings of the IEEE/CVF Conference on Computer Vision and Pattern
  Recognition}}. \bibinfo{pages}{10734--10742}.
\newblock


\bibitem[\protect\citeauthoryear{Wu, Wang, Zhang, Tian, Vajda, and Keutzer}{Wu
  et~al\mbox{.}}{2018}]%
        {wu2018mixed}
\bibfield{author}{\bibinfo{person}{Bichen Wu}, \bibinfo{person}{Yanghan Wang},
  \bibinfo{person}{Peizhao Zhang}, \bibinfo{person}{Yuandong Tian},
  \bibinfo{person}{Peter Vajda}, {and} \bibinfo{person}{Kurt Keutzer}.}
  \bibinfo{year}{2018}\natexlab{}.
\newblock \showarticletitle{Mixed precision quantization of convnets via
  differentiable neural architecture search}.
\newblock \bibinfo{journal}{{\em arXiv preprint arXiv:1812.00090\/}}
  (\bibinfo{year}{2018}).
\newblock


\bibitem[\protect\citeauthoryear{Xiao, Rasul, and Vollgraf}{Xiao
  et~al\mbox{.}}{2017}]%
        {xiao2017fashion}
\bibfield{author}{\bibinfo{person}{Han Xiao}, \bibinfo{person}{Kashif Rasul},
  {and} \bibinfo{person}{Roland Vollgraf}.} \bibinfo{year}{2017}\natexlab{}.
\newblock \showarticletitle{Fashion-mnist: a novel image dataset for
  benchmarking machine learning algorithms}.
\newblock \bibinfo{journal}{{\em arXiv preprint arXiv:1708.07747\/}}
  (\bibinfo{year}{2017}).
\newblock


\bibitem[\protect\citeauthoryear{Ying, Feng, Zhao, Dhar, Dalir, Gu, Cheng,
  Soref, Pan, and Chen}{Ying et~al\mbox{.}}{2020}]%
        {ying2020electronic}
\bibfield{author}{\bibinfo{person}{Zhoufeng Ying}, \bibinfo{person}{Chenghao
  Feng}, \bibinfo{person}{Zheng Zhao}, \bibinfo{person}{Shounak Dhar},
  \bibinfo{person}{Hamed Dalir}, \bibinfo{person}{Jiaqi Gu},
  \bibinfo{person}{Yue Cheng}, \bibinfo{person}{Richard Soref},
  \bibinfo{person}{David~Z Pan}, {and} \bibinfo{person}{Ray~T Chen}.}
  \bibinfo{year}{2020}\natexlab{}.
\newblock \showarticletitle{Electronic-photonic arithmetic logic unit for
  high-speed computing}.
\newblock \bibinfo{journal}{{\em Nature communications\/}}
  \bibinfo{volume}{11}, \bibinfo{number}{1} (\bibinfo{year}{2020}),
  \bibinfo{pages}{1--9}.
\newblock


\bibitem[\protect\citeauthoryear{Zhou, Lin, Wu, Chen, Xie, Li, Fan, Wu, Fang,
  and Dai}{Zhou et~al\mbox{.}}{2021}]%
        {zhou2021large}
\bibfield{author}{\bibinfo{person}{Tiankuang Zhou}, \bibinfo{person}{Xing Lin},
  \bibinfo{person}{Jiamin Wu}, \bibinfo{person}{Yitong Chen},
  \bibinfo{person}{Hao Xie}, \bibinfo{person}{Yipeng Li},
  \bibinfo{person}{Jingtao Fan}, \bibinfo{person}{Huaqiang Wu},
  \bibinfo{person}{Lu Fang}, {and} \bibinfo{person}{Qionghai Dai}.}
  \bibinfo{year}{2021}\natexlab{}.
\newblock \showarticletitle{Large-scale neuromorphic optoelectronic computing
  with a reconfigurable diffractive processing unit}.
\newblock \bibinfo{journal}{{\em Nature Photonics\/}} \bibinfo{volume}{15},
  \bibinfo{number}{5} (\bibinfo{year}{2021}), \bibinfo{pages}{367--373}.
\newblock


\end{thebibliography}

%\clearpage
 
%\onecolumn

%\appendix
%\input{sec-appendix}

\end{document}